\documentclass[sigconf, screen, balance=true]{acmart}
\AtBeginDocument{%
  \providecommand\BibTeX{{%
    \normalfont B\kern-0.5em{\scshape i\kern-0.25em b}\kern-0.8em\TeX}}}

\copyrightyear{2022}
\acmYear{2022}
\setcopyright{rightsretained}

\acmConference[KDD '22]{Proceedings of the 28th ACM SIGKDD Conference on Knowledge Discovery and Data Mining}{August 14--18, 2022}{Washington, DC, USA}
\acmBooktitle{Proceedings of the 28th ACM SIGKDD Conference on Knowledge Discovery and Data Mining (KDD '22), August 14--18, 2022, Washington, DC, USA}
\acmDOI{10.1145/3534678.3539074}
\acmISBN{978-1-4503-9385-0/22/08}

\usepackage{etoolbox}

\makeatletter

\newcommand\xlabel[2][]{\phantomsection\def\@currentlabelname{#1}\label{#2}}

\patchcmd{\maketitle}{\@copyrightpermission}{
   \begin{minipage}{0.3\columnwidth}
     \href{https://creativecommons.org/licenses/by/4.0/}{\includegraphics[width=0.90\textwidth]{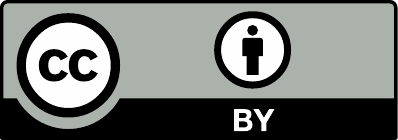}}
   \end{minipage}\hfill
   \begin{minipage}{0.7\columnwidth}
     \href{https://creativecommons.org/licenses/by/4.0/}{This work is licensed under a Creative Commons Attribution International 4.0 License.}
   \end{minipage}
   \vspace{5pt}
}{}{}
\makeatother

\usepackage{caption}
\usepackage{enumitem}
\usepackage{tabu}
\usepackage{rotating}
\usepackage{tikz}
\usepackage{capt-of,etoolbox}
\usepackage{microtype}
\usepackage{bm}
\usepackage{soul}
\usepackage{wrapfig}
\usepackage{graphbox}
\usepackage{tcolorbox}
\usepackage{cleveref}
\usepackage{mathtools}
\usepackage{balance}
\usepackage[varqu]{zi4}
\usepackage[bb=boondox]{mathalfa}
\usepackage[utf8x]{inputenc}
\usepackage{units}
\usepackage{setspace}
\usepackage{gensymb}
\usepackage{printlen} %
\usepackage{amsmath}
\usepackage{eqnarray}
\usepackage{bbm}

\definecolor{orange}{RGB}{250,130,49}
\definecolor{red}{RGB}{234,59,90}
\definecolor{agreen}{RGB}{74, 198, 148}
\definecolor{purple}{RGB}{158, 62, 177}
\definecolor{darkpurple}{RGB}{170, 70, 210}
\definecolor{aqua}{RGB}{87, 180, 181}
\definecolor{lightblue}{RGB}{72, 123, 232}
\definecolor{hotpink}{RGB}{255, 83, 115}
\definecolor{teal}{RGB}{90, 200, 250}
\definecolor{linkColor}{RGB}{0, 128, 229}
\definecolor{lightgreen}{RGB}{33, 222, 128}
\definecolor{gray}{RGB}{75, 101, 132}

\definecolor{myred}{RGB}{224, 49, 119}
\definecolor{myorange}{RGB}{250, 130, 49}
\definecolor{myyellow}{RGB}{254, 211, 48}
\definecolor{mygreen}{RGB}{14, 152, 136}
\definecolor{myblue}{RGB}{0, 128, 229}
\definecolor{myviolet}{RGB}{56, 103, 214}
\definecolor{mypurple}{RGB}{136, 84, 208}
\definecolor{mybrown}{RGB}{132, 99, 88}
\definecolor{mygray}{RGB}{220, 220, 220}

\definecolor{myACMBlue}{cmyk}{1,0.1,0,0.1}
\definecolor{myACMYellow}{cmyk}{0,0.16,1,0}
\definecolor{myACMOrange}{cmyk}{0,0.42,1,0.01}
\definecolor{myACMRed}{cmyk}{0,0.90,0.86,0}
\definecolor{myACMLightBlue}{cmyk}{0.49,0.01,0,0}
\definecolor{myACMGreen}{cmyk}{0.20,0,1,0.19}
\definecolor{myACMPurple}{cmyk}{0.55,1,0,0.15}
\definecolor{myACMDarkBlue}{cmyk}{1,0.58,0,0.21}

\definecolor{refColor}{RGB}{109, 35, 130}

\newcommand{\link}[1]{{\href{#1}{\color{myACMDarkBlue}\textbf{\texttt{#1}}}}}
\newcommand{\figpart}[1]{\textcolor{refColor}{#1}}

\newcommand{\mcolor}[2]{\textcolor{#1}{#2}}
\newcommand{\tcolor}[2]{\textcolor{#1}{#2}}

\crefname{figure}{fig.}{fig.}
\Crefname{figure}{Fig.}{Fig.}
\crefname{equation}{eq.}{eq.}
\Crefname{equation}{Eq.}{Eq.}
\crefname{section}{\S}{\S}
\creflabelformat{equation}{#2\textup{#1}#3}

\newcommand{\tool}{\textsc{\textsf{GAM Changer}}}

\newtoggle{inheader}
\newcommand{\canvasview}{\iftoggle{inheader}{GAM Canvas}{\textit{GAM Canvas}}}
\newcommand{\metricview}{\iftoggle{inheader}{Metric Panel}{\textit{Metric Panel}}}
\newcommand{\featureview}{\iftoggle{inheader}{Feature Panel}{\textit{Feature Panel}}}
\newcommand{\historyview}{\iftoggle{inheader}{History Panel}{\textit{History Panel}}}
\newcommand{\toolbar}{\iftoggle{inheader}{Context Toolbar}{\textit{Context Toolbar}}}
\newcommand{\statusbar}{\iftoggle{inheader}{Status Bar}{\textit{Status Bar}}}

\definecolor{soulorange}{RGB}{255, 212, 153}
\definecolor{soulgray}{RGB}{220, 220, 220}
\definecolor{soulgraylight}{RGB}{235, 235, 235}
\definecolor{soulred}{RGB}{252, 217, 218}
\definecolor{soulbluelight}{RGB}{208, 233, 253}
\definecolor{souldorangelight}{RGB}{254, 234, 212}
\colorlet{soulblue}{myblue!30}

\newcommand{\bluelighthl}[1]{{\sethlcolor{soulbluelight}\hl{#1}}}

\newcommand{\orangelighthl}[1]{{\sethlcolor{souldorangelight}\hl{#1}}}

\newcommand{\graylighthl}[1]{{\sethlcolor{soulgraylight}\hl{#1}}}

\newcommand{\inlinefig}[2]{\protect\includegraphics[align=c, height=#1pt]{figures/#2}}

\definecolor{tagbordercolor}{rgb}{0.8, 0.8, 0.8}
\definecolor{tagbgcolor}{rgb}{0.9, 0.9, 0.9}

\newtcbox{\tagg}{nobeforeafter, colframe=tagbordercolor,
colback=tagbgcolor, boxrule=0.5pt, arc=1pt,
  boxsep=0pt,left=2pt,right=2pt,top=1.5pt,bottom=2pt,tcbox raise base}

\definecolor{lightgray}{RGB}{247, 247, 247}
\definecolor{midgray}{RGB}{179, 179, 179}

\newtcbox{\featuretag}{on line,
  colframe=midgray,colback=lightgray,
  boxrule=0.5pt,arc=2pt,boxsep=0pt,left=2pt,right=1pt,top=1pt,bottom=1pt
}
\newcommand*\myquote[1]{``\textit{#1}''}

\definecolor{tagbgcolor}{rgb}{1, 1, 1}

\definecolor{boxyellow}{RGB}{206, 171, 1}
\definecolor{boxgreen}{RGB}{14, 152, 136}
\definecolor{boxblue}{RGB}{77, 167, 223}

\setul{0.45ex}{0.2ex}

\newcommand*\mybox[1]{\setulcolor{#1}\ul}

\newcommand{\headermargintop}{\vspace{-4px}}
\newcommand{\headermarginbottom}{\vspace{-0.6px}}

\begin{document}

\title{Interpretability, Then What? Editing Machine Learning Models to Reflect Human Knowledge and Values}

\newcommand{\authorgap}{\hspace{3pt}}

\author{Zijie J. Wang}
\orcid{0000-0003-4360-1423}
\affiliation{%
  \institution{Georgia Institute of Technology}
  \country{}
}

\author{Alex Kale}
\orcid{0000-0001-7668-2800}
\affiliation{%
  \institution{University of Washington}
  \country{}
}

\author{Harsha Nori}
\orcid{0000-0002-5442-1359}
\affiliation{%
  \institution{Microsoft Research}
  \country{}
}

\author{Peter Stella}
\orcid{0000-0001-7521-2516}
\author{Mark E. Nunnally}
\orcid{0000-0003-2129-4733}
\affiliation{%
  \institution{NYU Langone Health}
  \country{}
}

\author{Duen Horng Chau}
\orcid{0000-0001-9824-3323}
\affiliation{%
  \institution{Georgia Institute of Technology}
  \country{}
}

\author{Mihaela Vorvoreanu}
\orcid{0000-0002-3322-3548}
\author{Jennifer Wortman Vaughan}
\orcid{0000-0002-7807-2018}
\author{Rich Caruana}
\orcid{0000-0002-6383-7786}
\affiliation{%
  \institution{Microsoft Research}
  \country{}
}

\renewcommand{\shortauthors}{Zijie J. Wang et al.}

\begin{abstract}
  Machine learning (ML) interpretability techniques can reveal undesirable patterns in data that models exploit to make predictions---potentially causing harms once deployed.
  However, how to take action to address these patterns is not always clear.
  In a collaboration between ML and human-computer interaction researchers, physicians, and data scientists, we develop \tool{}, the first interactive system to help domain experts and data scientists easily and responsibly edit Generalized Additive Models (GAMs) and fix problematic patterns.
  With novel interaction techniques, our tool puts interpretability into action---empowering users to analyze, validate, and align model behaviors with their knowledge and values.
  Physicians have started to use our tool to investigate and fix pneumonia and sepsis risk prediction models, and
  an evaluation with 7 data scientists working in diverse domains highlights that our tool is easy to use, meets their model editing needs, and fits into their current workflows.
  Built with modern web technologies, our tool runs locally in users' web browsers or computational notebooks, lowering the barrier to use.
  \tool{} is available at the following public demo link: \link{https://interpret.ml/gam-changer}.
\end{abstract}




\maketitle
\vspace{-2px}
\section{Introduction}
\vspace{-2px}

\setlength{\belowcaptionskip}{-11.3pt}
\setlength{\abovecaptionskip}{3pt}
\begin{figure}[tb]
  \vspace{3pt}
  \includegraphics[width=\linewidth]{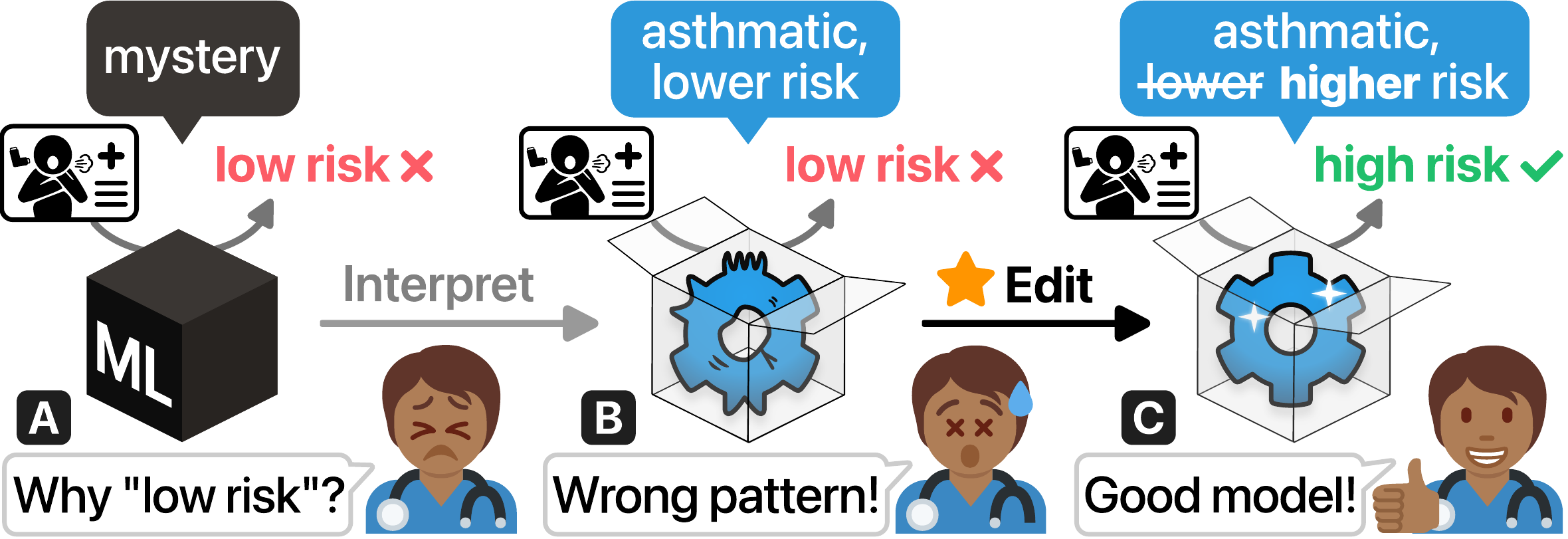}
  \Description{Motivation for the study}
  \caption{
    (A) Domain experts such as physicians often hesitate to trust ML models as they cannot understand how the models make predictions.
    (B) Interpretability reveals models can learn potentially harmful patterns.
    (C) Model editing turns interpretability into action---enabling domain experts to align model behaviors with their knowledge and values.\looseness=-1
  }
  \label{fig:motivation}
\end{figure}
\setlength{\belowcaptionskip}{0pt}
\setlength{\abovecaptionskip}{12pt}

It is crucial to understand how machine learning (ML) models make predictions in high-stakes settings, such as finance, criminal justice, and healthcare~(\autoref{fig:motivation}\figpart{A}).
Recently, researchers have made substantial efforts to make ML models interpretable~\cite[e.g.,][]{ribeiroWhyShouldTrust2016,lundbergUnifiedApproachInterpreting2017,caruanaIntelligibleModelsHealthCare2015}, but there is not much research focused on how to \textit{act on} model interpretations.
In practice, data scientists and domain experts often compare model interpretations with their knowledge~\cite{hongHumanFactorsModel2020}.
If a model uses expected patterns to make predictions, they feel more confident to deploy it.
Interpretability can also uncover hidden relationships in the data---helping users gain insights into the problems they want to tackle. \looseness=-1

Other times, however, ML interpretability reveals that models learn dangerous patterns from the data and rely on these patterns to make predictions.
These patterns might accurately reflect real phenomena, but leaving them untouched can cause serious harm in deployment.
For example, with interpretability, KDD researchers~\cite{caruanaIntelligibleModelsHealthCare2015, changHowInterpretableTrustworthy2021} found healthcare models predict that having asthma lowers a patient's risk of dying from pneumonia~(\autoref{fig:motivation}\figpart{B}). Researchers suspect this is because asthmatic patients would receive care earlier, leading to better outcomes in the training data.
If we use these flawed models to make hospital admission decisions, asthmatic patients are likely to miss out on care they need.
Interpretability helps us identify these dangerous patterns, but how can we take a step further and use model explanations to
\textit{improve}~(\autoref{fig:motivation}\figpart{C}) ML models?

\begin{figure*}[tb]
  \includegraphics[width=\textwidth]{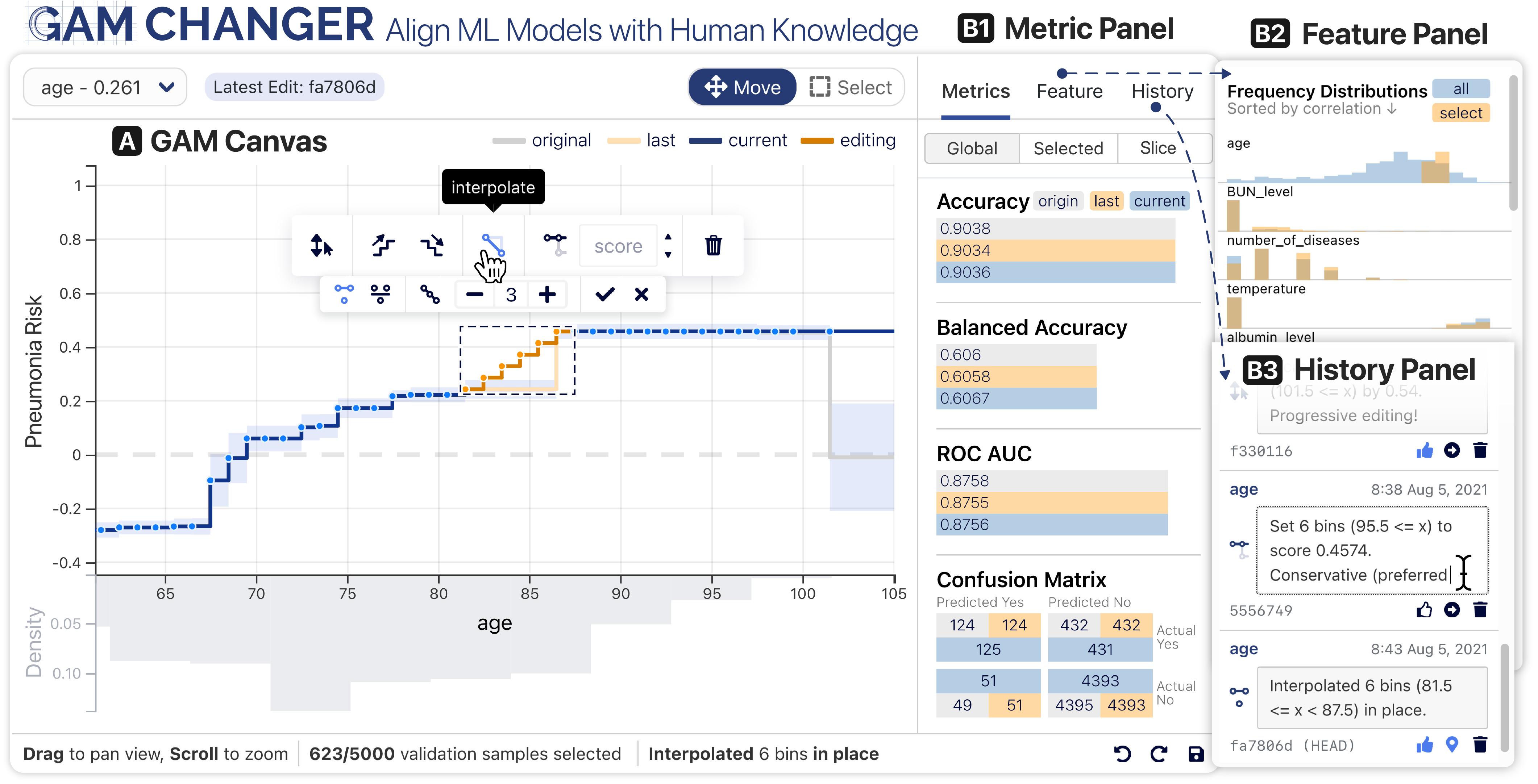}
  \Description{User interface for \tool{}}
  \caption{
    \tool{} empowers domain experts and data scientists to easily and responsibly
    align model behaviors with their knowledge and values, via direct manipulation of GAM model weights.
    Take a healthcare model for example.
    (A) The \canvasview{} enables physicians to interpolate the predicted risk of dying from pneumonia to match their clinical knowledge of a gradual risk increase from age \texttt{81} to age \texttt{87}.
    (B1) The \metricview{} provides real-time feedback on model performance.
    (B2) The \featureview{} helps users identify characteristics of affected samples and promotes awareness of fairness issues.
    (B3) The \historyview{} allows users to compare and revert changes, as well as document their motivations and editing contexts.
  }
  \label{fig:teaser}
\end{figure*}

To answer this question, our research team---consisting of ML and human-computer interaction (HCI) researchers, physicians, and data scientists---presents \textbf{\tool{}}~(\autoref{fig:teaser}): the first interactive system to empower domain experts and data scientists to easily and responsibly edit the weights of \textit{generalized additive models} (GAMs)~\cite{hastieGeneralizedAdditiveModels1999, louAccurateIntelligibleModels2013,louIntelligibleModelsClassification2012}, a state-of-the-art interpretable model~\cite{wangPursuitInterpretableFair2020}.
Model editing is already common practice for regulatory compliance~(\autoref{sec:study-needs}). We aim to tackle two critical challenges to make model editing more accessible and responsible:
\textbf{Challenge 1: Enable domain experts to vet and fix models.}\xlabel[Challenge 1]{item:c1}
Editing model weights to align model behavior with domain knowledge has been discussed in the KDD community~\cite{caruanaIntelligibleModelsHealthCare2015}.
It requires the ``editors'' to have expertise in ML engineering and write code to adjust specific weights until the model behaves as expected.
However, domain experts who have less experience in ML engineering, such as physicians and legal experts, play a critical role in creating trustworthy models~\cite{hongHumanFactorsModel2020}.
To democratize model editing, we develop easy-to-use and flexible user interfaces that support a wide range of editing methods---enabling stakeholders with diverse backgrounds to easily investigate and improve ML models.

\textbf{Challenge 2: Promote accountable model modifications.}\xlabel[Challenge 2]{item:c2}
Accessible model editing helps users exercise their human agency but demands caution, as modifications of high-stake models have serious consequences.
For example, if a user only monitors edits' effects on a metric like overall accuracy, their edits might have unfavorable effects on underrepresented groups~\cite{wallWarningBiasMay2017}.
To guard against harmful edits, we provide users with continuous feedback about impacts on different subgroups and feature correlations. We also support transparent and reversible model modifications.

\smallskip
\noindent
\textbf{Contributions \& Impacts.}
\tool{} has already begun to help users improve their models.
Our major contributions include:

\begin{itemize}[topsep=1mm, itemsep=0mm, parsep=1mm, leftmargin=3mm]

\item \textbf{\tool{}, the first interactive system} that empowers domain experts and data scientists to edit GAMs to align model behaviors with their knowledge and values.
Through a participatory and iterative design process with physicians and data scientists, we adapt easy-to-use \textit{direct manipulation}~\cite{shneidermanDirectManipulationStep1983} interfaces to edit complex ML models.
Guarding against harmful edits is our priority:
we employ \textit{continuous feedback} and \textit{reversible actions} to elucidate editing effects and promote accountable edits~(\autoref{sec:experience}).\looseness=-1

\item \textbf{Impacts to physicians: \tool{} in action.}
Physicians have started to use our tool to vet and fix healthcare ML models.
We present two examples where physicians on our team applied \tool{} to align pneumonia and sepsis risk predictions with their clinical knowledge.
The edited sepsis risk prediction model will be adapted for use in a large hospital~(\autoref{sec:case-study}).

\item \textbf{Impacts to data scientists: beyond healthcare.}
To investigate how our tool will help ML practitioners, we further evaluate it via a user study with 7 data scientists in finance, healthcare, and media.
Our study suggests \tool{} is easy to understand, fits into practitioners' workflow, and is especially enjoyable to use.
We also find model editing via feature engineering and parameter tuning is a common practice for regulatory compliance.
Reflecting on our study, we derive lessons and future directions for model editing and interpretability tools~(\autoref{sec:user-study}, \autoref{sec:discussion}).

\item \textbf{An open-source,\footnote{\tool{} code: \link{https://github.com/interpretml/gam-changer}} web-based implementation} that broadens people's access to creating more accountable ML models and exercising their human agency in a world penetrated by ML systems.
We develop \tool{} with modern web technologies such as WebAssembly.\footnote{WebAssembly: \link{https://webassembly.org}}
Therefore, anyone can access our tool directly in their web browser or computational notebooks and edit ML models with their own datasets at scale~(\autoref{sec:experience:implementation}).
For a demo video of \tool{}, visit \link{https://youtu.be/D6whtfInqTc}.

\end{itemize}

\noindent We hope our work helps emphasize the importance of human agency in responsible ML research, and inspires and informs future work in human-AI interaction and actionable ML interpretability.

\headermargintop{}
\section{Background \& Related Work}
\headermarginbottom{}

Generalized additive models (GAMs) have emerged as a popular model class among the data science community.
GAMs' predictive performance is on par with complex black-box models~\cite{wangPursuitInterpretableFair2020}, yet GAMs remain simple enough for humans to understand their decision process~\cite{changHowInterpretableTrustworthy2021}.
Given an \tcolor{myorange}{input} $\mcolor{myorange}{x \in \mathbb{R}^{k}}$ with \tcolor{myorange}{$k$ features} and a \tcolor{myred}{target} $\mcolor{myred}{y \in \mathbb{R}}$, a GAM can be written as:
\begin{equation}
    \label{equation:gam}
    \mcolor{gray}{g \left(\mcolor{myred}{y}\right)} = \mcolor{mygreen}{\beta_0} + \mcolor{myblue}{f_1 \left(\mcolor{myorange}{x_1}\right)} + \mcolor{myblue}{f_2 \left(\mcolor{myorange}{x_2}\right)} + \cdots + \mcolor{myblue}{f_k \left(\mcolor{myorange}{x_k}\right)}
\end{equation}
Different \tcolor{gray}{link functions} $\mcolor{gray}{g}$ are appropriate for different tasks: we use the \tcolor{gray}{logit} for binary classification and \tcolor{gray}{identity} for regression.
$\mcolor{mygreen}{\beta_0}$ is \tcolor{mygreen}{the intercept}.
There are many options for \tcolor{myblue}{shape functions} $\mcolor{myblue}{f_j}$, such as \tcolor{myblue}{splines}~\cite{hastieGeneralizedAdditiveModels1999}, \tcolor{myblue}{gradient-boosted trees}~\cite{caruanaIntelligibleModelsHealthCare2015}, and \tcolor{myblue}{neural networks}~\cite{agarwalNeuralAdditiveModels2021}.
Some GAMs support pair-wise interaction terms $\mcolor{myblue}{f_{ij}(\mcolor{myorange}{x_i, x_j})}$.

GAMs are interpretable and editable because people can visualize and modify each \tcolor{myorange}{feature $x_j$}'s contribution to the model's prediction by inspecting and adjusting the \tcolor{myblue}{shape function} $\mcolor{myblue}{f_j}$;
\tcolor{myblue}{$f_j(\mcolor{myorange}{x_j})$}  is sometimes called the feature's contribution score.
Since GAMs are additive, we can edit different \tcolor{myblue}{shape functions} independently.

\textbf{Model interpretability.}
Besides \textit{glass-box} models like GAMs that are inherently interpretable~\cite[e.g.,][]{zengInterpretableClassificationModels2017,lakkarajuInterpretableDecisionSets2016}, ML researchers have developed post hoc explanation methods to interpret \textit{black-box} models~\cite[e.g.,][]{ribeiroWhyShouldTrust2016,lundbergUnifiedApproachInterpreting2017} and have studied how interpretability methods are understood and used~\cite[e.g.,][]{kaurInterpretingInterpretabilityUnderstanding2020,hongHumanFactorsModel2020, wexlerWhatIfToolInteractive2019}. Closely related to our work, researchers have developed visualization tools specific to GAMs like \textsc{mgcViz}~\cite{fasioloScalableVisualizationMethods2020}, \textsc{Gamut}~\cite{hohmanGamutDesignProbe2019}, and \textsc{TeleGam}~\cite{hohmanTeleGamCombiningVisualization2019}.
Our work advances the interpretable ML landscape in making interpretability \textit{actionable} by enabling users to interactively fix their models.

\textbf{Model modification.}
Although being able to modify models leads to greater trust and better human-AI team performance~\cite{dietvorstOvercomingAlgorithmAversion2018}, research in model modification is relatively nascent.
By manipulating certain important neurons, researchers can modify a few semantic concepts in generated images~\cite{bauUnderstandingRoleIndividual2020}, control some text translation styles~\cite{bauIdentifyingControllingImportant2019}, and induce basic concepts in text generation~\cite{suauFindingExpertsTransformer2020}.
However, these works rely on post hoc explanations---users can only affect a small subset of model behaviors, and modifications are likely to have unknown effects.
Grounded in accurate and complete interpretations from glass-box models, \tool{} is the first system that empowers users to have total control of their model's behavior and observe full editing effects---enabling them to easily and safely improve ML models in potentially high-stakes settings.

\setlength{\belowcaptionskip}{-3.15pt}
\setlength{\abovecaptionskip}{5pt}
\setlength{\belowcaptionskip}{0pt}
\setlength{\abovecaptionskip}{12pt}

\headermargintop{}
\section{Novel User Experience}
\label{sec:experience}
\headermarginbottom{}

To lower barriers to controlling ML model behavior~\mbox{(\nameref{item:c1})}, \tool{}~(\autoref{fig:teaser}) adapts easy-to-use direct manipulation interface patterns to edit the parameters of GAMs with a variety of editing tools~(\autoref{sec:experience:method}).
To promote responsible edits~(\nameref{item:c2}), our tool provides real-time feedback; all edits are reversible, and users can document and compare their edits~(\autoref{sec:experience:responsible}).
Built with modern web technologies, our tool is accessible and scalable~(\autoref{sec:experience:implementation}).

\headermargintop{}
\subsection{Intuitive and Flexible Editing}
\label{sec:experience:method}
\headermarginbottom{}

The \canvasview{}~(\autoref{fig:teaser}\figpart{A}) is the main view of \tool{}, where we visualize one \tcolor{myorange}{input feature $\mcolor{myorange}{x_j}$}'s contribution to the model's prediction by plotting its \tcolor{myblue}{shape function} $\mcolor{myblue}{f_j(} \mcolor{myorange}{x_j} \mcolor{myblue}{)}$.
Users can select a drop-down to transition across features.
GAMs usually discretize continuous variables into finite bins, so that shape functions can easily capture complex non-linear relationships.
Thus, the output of shape functions is a continuous piecewise constant function, where we use a dot to show the start of each bin and a line to encode the bin's constant score~(\autoref{fig:teaser}\figpart{A}).
For categorical features, we represent each bin as a bar whose height encodes the bin's score~(\autoref{fig:gallery}\figpart{B}).
Lines and bins are colored by editing status (e.g., original or edited).

\textbf{Model direct manipulation.}
In the \canvasview{}, users can \textit{zoom-and-pan} to control their viewpoint in the \textit{move mode}, or use \textit{marquee selection} to select a region of the shape function to edit in the \textit{select mode}~(\autoref{fig:teaser}\figpart{A}).
Once a region is selected, the \toolbar{} appears: it affords a variety of editing tools represented as icon buttons.
Clicking a button changes the shape function in the selected region.
For example, the monotonicity tool~\inlinefig{10}{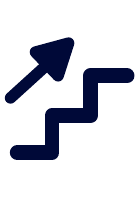} can transform the selected interval of the shape function into a monotonically increasing function.
Internally, \tool{} fits an isotonic regression~\cite{barlowStatisticalInferenceOrder1972} weighted by the bin counts to determine a monotone function with minimal changes.
Other editing tools include interpolating~\inlinefig{10}{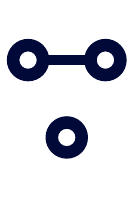} scores of selected bins, dragging~\inlinefig{10}{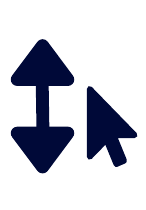} to adjust scores, and aligning~\inlinefig{10}{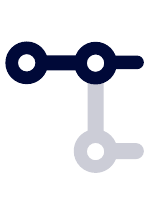} scores to the most left or right bin~(see \nameref{appendix:tools} for details).

\headermargintop{}
\subsection{Safe and Responsible Editing}
\label{sec:experience:responsible}
\headermarginbottom{}
Guarding against harmful edits is our top priority.
To begin using \tool{}, users provide a trained GAM (i.e., model weights) and set of validation samples (a subset of the training data or separate validation set).
The \metricview{}~(\autoref{fig:teaser}\figpart{-B1}) provides real-time and continuous feedback on the model's performance on the validation samples to help users identify the effects of their edits.
During a user's editing process, our tool efficiently recomputes performance metrics on the edited model.
To probe if an edit is equitable across different subgroups, users can choose which subset of samples to measure performance on: the \textit{Global Scope} for all samples, the \textit{Selected Scope} for samples in the selected region, and the \textit{Slice Scope} for samples having a specific categorical value (e.g., females).

\textbf{Recognize impact disparities.}
The \featureview{}~(\autoref{fig:teaser}\figpart{-B2}) helps users gain an overview of correlated features and elucidates potential disparities in the impact of edits.
For example, it can alert users of the disproportionate impact of edits addressing elder patients on females as females live longer.
We develop \textit{linking+reordering}---a novel method to identify correlated features.
Once a user selects a region in the \canvasview{}, we look up affected samples' associated bins across all features.
For each feature, we compute the $\ell_2$ distance between the bin count frequency in all training data and the frequency in affected samples.
By observing overlaid histograms sorted in descending order of the distance scores, users can inspect correlated features of affected samples and identify potential editing effect disparity (see \nameref{sec:featureview} for details).

\xlabel[]{sec:experience:history}
\textbf{Reversible and documented modifications.}
To promote safe model editing, \tool{} allows users to undo and redo any edits (see \nameref{sec:historyview} for details).
In addition, the \historyview{}~(\autoref{fig:teaser}\figpart{-B3}) tracks all edits and displays each edit in a list.
Inspired by the version control system Git, we save each edit as a commit---a snapshot of the underlying GAM weights.
Each commit has a timestamp, a unique identifier, and a commit message.
Therefore, users can easily explore model evolution by checking out~\inlinefig{10}{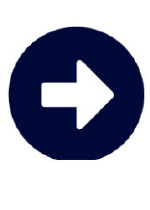} a previous GAM version, discard~\inlinefig{9}{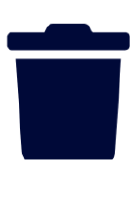} modifications, and document edit contexts and motivations in commit messages.
Once satisfied with their edits, a user can save the modified model with edit history for deployment or future continuing editing.
To help users identify editing mistakes and promote accountable edits, \tool{} requires users to examine and confirm~\inlinefig{10}{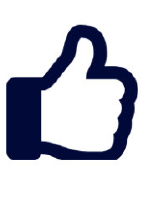} all edits before saving the model.

\subsection{Scalable, Open-source Implementation}
\label{sec:experience:implementation}
\tool{} is a web-based GAM editor that users can access with any web-browsers on laptops or tablets, or directly in computational notebooks.
Our tool has been integrated into the popular ML interpretability ecosystem \textit{InterpretML}~\cite{noriInterpretMLUnifiedFramework2019}: users can easily \textit{export} models to edit and \textit{load} modified models.
Using cutting-edge \mbox{WebAssembly} to accelerate in-browser model inference and isotonic regression fitting, our tool is scalable: all computations are real-time with up to 5k validation samples in Firefox on a MacBook, and the sample size is only bounded by the browser's memory limit.
We open source \tool{} so that future researchers can easily generalize our design to other forms of model editing.\looseness=-1 %
\section{Impacts to physicians}
\label{sec:case-study}

\textbf{\tool{} in action.}
The early prototype~\cite{wangGAMChangerEditing2021} of our tool has received overwhelmingly positive feedback in two physician-focused workshops.\footnote{We presented our work-in-progress in the American Association of Physicists in Medicine Practical Big Data Workshop, and the NeurIPS Workshop on Bridging the Gap: From Machine Learning Research to Clinical Practice, both without proceedings.}
In addition, physicians have begun to use our tool to interpret and edit medical models.
We share examples in which two physicians in our research team apply \tool{} to investigate and improve GAMs for sepsis~(\autoref{sec:case-study-2}) and pneumonia~(\autoref{sec:case-study-1}) risk predictions, editing the models to reflect their clinical knowledge and values such as safety.
The edited sepsis risk prediction model will be adapted for use in a large hospital.

\subsection{Fixing Sepsis Risk Prediction}
\label{sec:case-study-2}
A physician in our team trained a GAM with boosted-trees to predict if pediatric patients should receive sepsis treatments.
This model exhibited many problematic patterns.
In this section, we share our experience in applying \tool{} to align this model's behavior with the physician's clinical knowledge and values.

The data comes from a large hospital; it includes 26,564 pediatric patients.
There are 7 continuous features: \inlinefig{9}{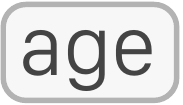}, \inlinefig{9}{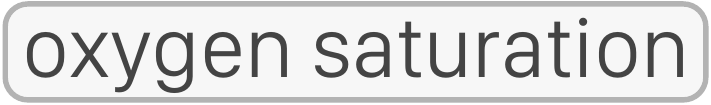}, \inlinefig{9}{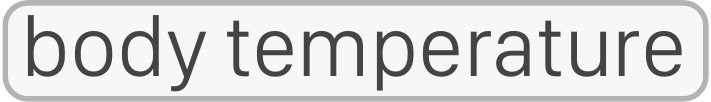}, systolic and diastolic~\inlinefig{9}{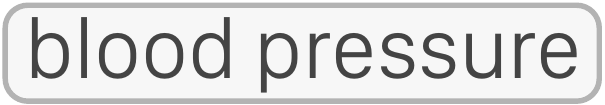}, \inlinefig{9}{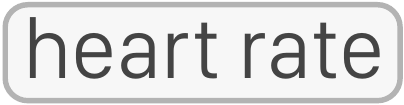}, and \inlinefig{9}{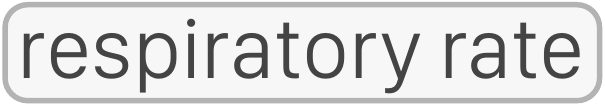}.
The \inlinefig{9}{tag-pressure}, \inlinefig{9}{tag-heart}, and \inlinefig{9}{tag-breath} are normalized by taking the difference between the original value and the age-adjusted normal.
The other 83 features are categorical with binary values, each indicating if a keyword---such as ``pain,'' ``fever,'' or ``fall''---is present in the \textit{chief complaint of patient}, a concise statement describing the symptom, diagnosis, and other reasons for a medical encounter.
The target variable is binary: \texttt{1} if the patient received a treatment for sepsis and \texttt{0} if not.
The model yields an AUC score of 0.865 on the test set (20\% of all data).
The physician loads \tool{} in their browser with 5,000 random training samples; they share their computer screen with 3 other researchers in the team via video-conferencing software.
All edits are made by the physician after discussing with other researchers on the call.

\setlength{\belowcaptionskip}{-8pt}
\setlength{\abovecaptionskip}{5pt}
\begin{figure}[tb]
  \includegraphics[width=\linewidth]{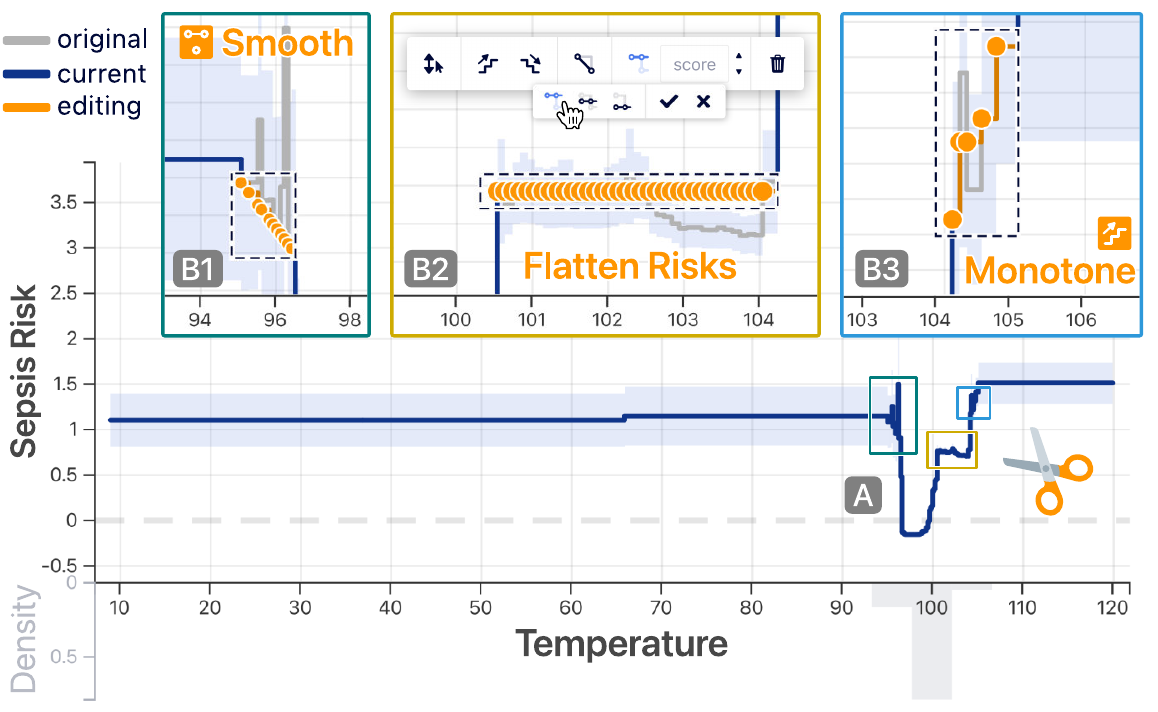}
  \Description{User interface for \tool{}}
  \caption{
    \inlinefig{9}{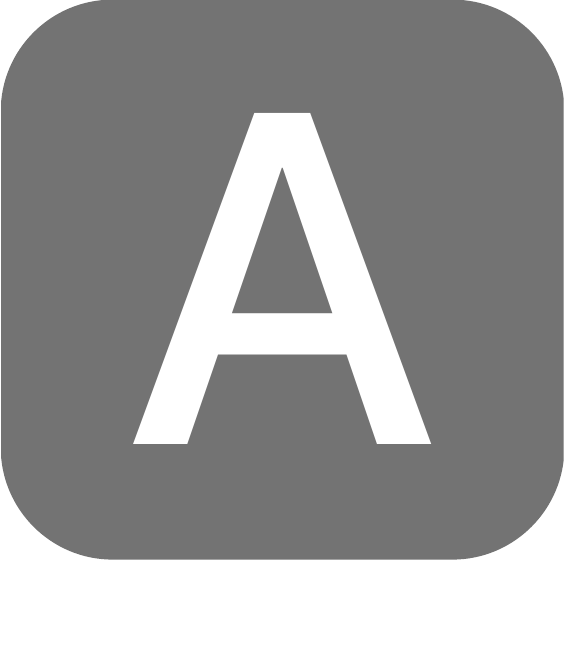} A GAM learns a few strange patterns between patients' temperature and sepsis risk that need to be fixed.
    \inlinefig{9}{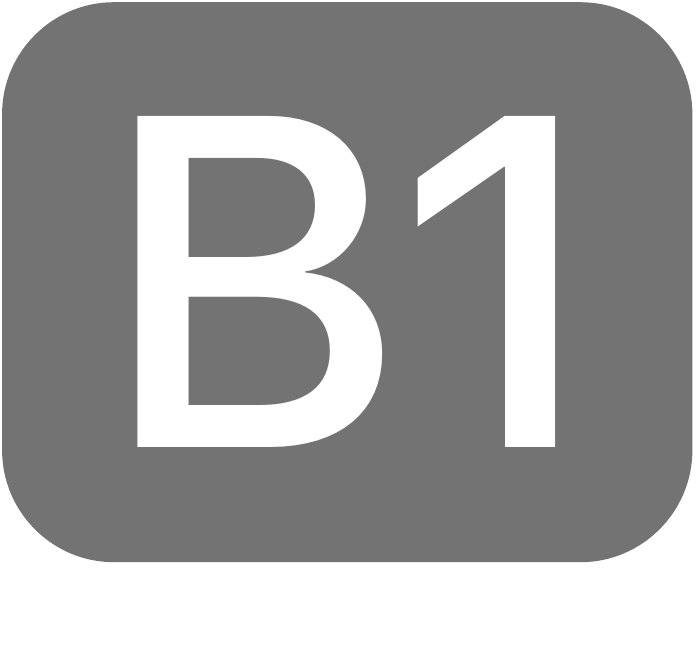}~We smooth out the sudden increase of risk~\inlinefig{10}{interpolate} around \texttt{96}$^{\circ}$F, \inlinefig{9}{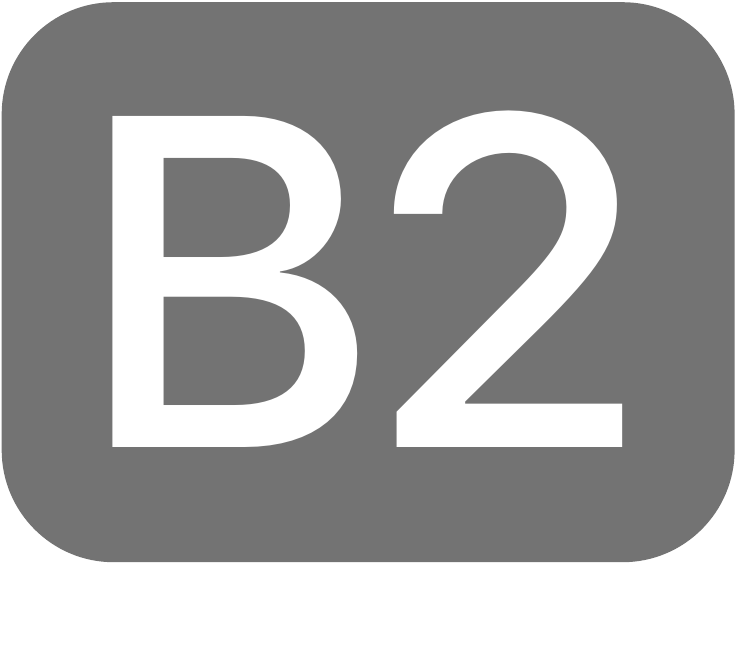}~flatten the risk~\inlinefig{10}{align} to reflect a treatment effect, and \inlinefig{10}{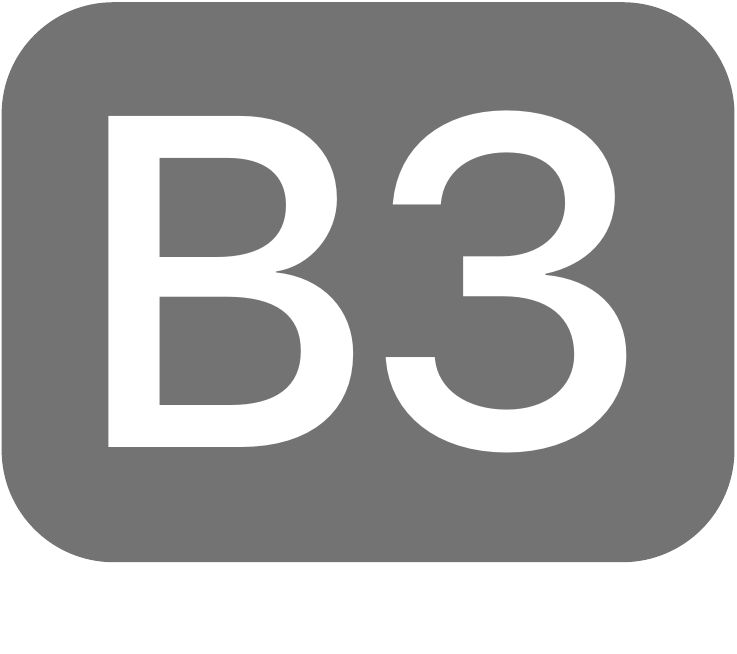} smooth out risk fluctuations~\inlinefig{10}{increasing} at high temperature.
  }
  \label{fig:temperature}
\end{figure}
\setlength{\belowcaptionskip}{0pt}
\setlength{\abovecaptionskip}{12pt}

\subsubsection{Editing the temperature feature.}
The \canvasview{} first shows \inlinefig{9}{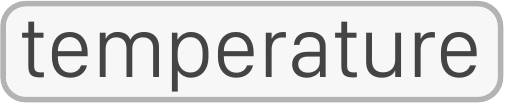}~(\autoref{fig:temperature}\figpart{A}) since this feature has the highest importance score, computed as the weighted average of a feature's absolute contribution scores.
The x-axis ranges from \texttt{10} to \texttt{120}$^{\circ}$F, where the low range is due to data errors.
The y-axis encodes the predicted risk score (log odds) of dying from sepsis, ranging from \texttt{-0.2} to \texttt{1.5}.
The shape function has a ``U-shape'': the model predicts that patients with \inlinefig{9}{tag-temperature} lower and higher than the normal range (\texttt{97}--\texttt{99}$^{\circ}$F) have a higher risk of sepsis.
It matches clinical knowledge as fever (high \inlinefig{9}{tag-temperature}) and hypothermia (low \inlinefig{9}{tag-temperature} caused by cardiovascular collapse) are severe symptoms of sepsis.
There is a peak of predicted risk when the \inlinefig{9}{tag-temperature} is \mybox{boxgreen}{around 96$^{\circ}$F}.
However, there is no physiological reason that hypothermia with a \inlinefig{9}{tag-temperature} of \texttt{96}$^{\circ}$F has a higher risk than a \inlinefig{9}{tag-temperature} of \texttt{95}$^{\circ}$F.
Therefore, we remove the risk peak at \texttt{96}$^{\circ}$F by linearly interpolating~\inlinefig{9}{interpolate} the risk scores from \texttt{95} to \texttt{96.5}$^{\circ}$F~(\autoref{fig:temperature}\figpart{-B1}).\looseness=-1

There is a plateau of risk scores \mybox{boxyellow}{from \texttt{100}\textnormal{--}\texttt{104}$^{\circ}$F}, with a small, but notable dip from \texttt{103}--\texttt{104}$^{\circ}$F.
The presence of the plateau itself is physiologically plausible (due to antipyretic treatments), but the dip is hard to explain and suspicious, perhaps reflecting a treatment effect in which treatment is delayed outside of the model’s prediction window as physicians evaluate the child’s response to antipyretics.
Because of a concern that this might artificially depress risk scores and encourage physicians to believe that children in this range are healthier than they really are, the risk curve in this region is flattened using the align tool~\inlinefig{9}{align}~(\autoref{fig:temperature}\figpart{-B2}).

Similarly, the observation of many small dips of predicted risk scores \mybox{boxblue}{around \texttt{104}\textnormal{--}\texttt{105.5}$^{\circ}$F} does not align with physiological knowledge.
Therefore, we remove these dips by making the scores monotonically increasing~\inlinefig{9}{increasing} in this region by fitting an isotonic regression model.
The physician in our team thinks this edit is conservative and safe because it smooths out many dips in the region that might cause patients to lose necessary care.
The physician comments \myquote{Taking out unpredictable behaviors from a model to my mind is deeply safer. If this ends up being a life and death decision, and we go back, and we look that a kid died because he didn’t trigger the model by falling into one of those dips, then that is a catastrophe.}

\setlength{\belowcaptionskip}{-4.7pt}
\setlength{\abovecaptionskip}{7pt}
\begin{figure}[tb]
  \includegraphics[width=\linewidth]{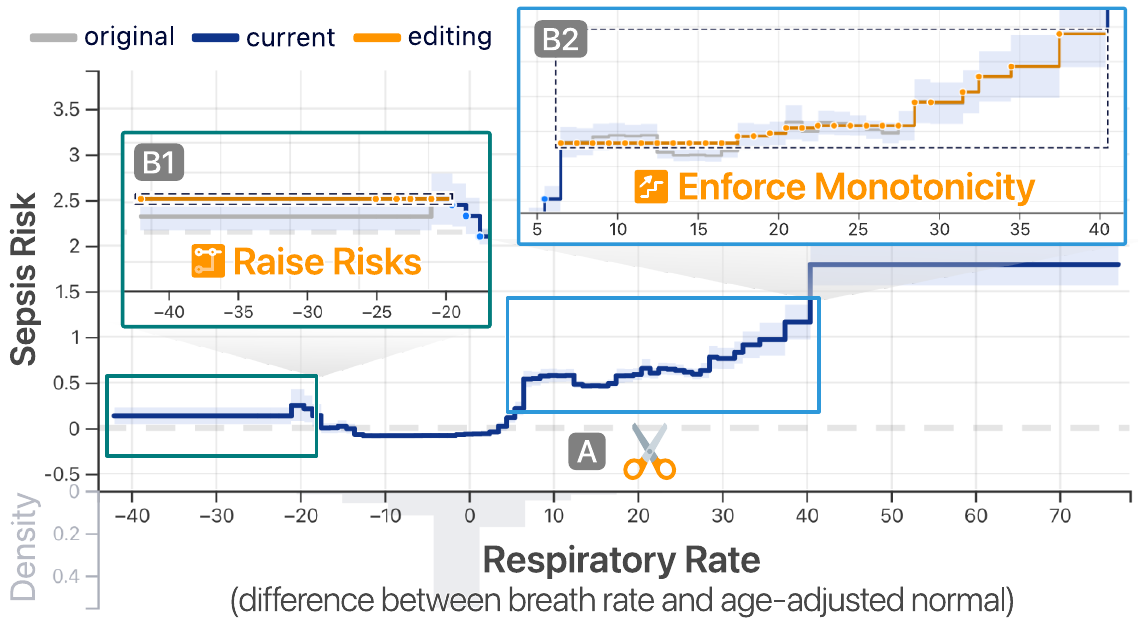}
  \Description{User interface for \tool{}}
  \caption{
    \inlinefig{9}{a3} Contrary to clinical knowledge, a GAM predicts sepsis risk decreases when the respiratory rate decreases~(left), and the risk score fluctuates when the rate increases~(right).
    We align the model behaviors by \inlinefig{9}{b3-1} raising risk scores~\inlinefig{10}{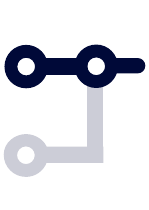} and \inlinefig{9}{b3-2} removing risk fluctuations with monotonicity~\inlinefig{10}{increasing}.\looseness=-2
  }
  \label{fig:breath}
\end{figure}
\setlength{\belowcaptionskip}{0pt}
\setlength{\abovecaptionskip}{12pt}

\subsubsection{Editing the respiratory rate feature.}
The \inlinefig{9}{tag-breath} feature measures the \textit{difference} between the number of breaths taken in one minute and its age-adjusted normal.
The ``U-shape'' in the \canvasview{}~(\autoref{fig:breath}\figpart{A}) suggests the model predicts that patients with high deviation from the normal respiratory rate range have a higher risk of sepsis, and higher \inlinefig{9}{tag-breath} yields a higher risk score than lower \inlinefig{9}{tag-breath}.
This pattern matches the clinical knowledge.
Interestingly, the center of the ``U-shape'' is around \texttt{-5} instead of \texttt{0}.
This also makes sense because the ``normal range'' of respiratory rate for adults is considered \texttt{12}--\texttt{20} times a minute, but healthy adults actually only take \texttt{12}--\texttt{15} breaths per minute.
In other words, this left-shifted center indicates the model has learned a realistic distribution of respiratory rate.

The predicted risk decreases when \inlinefig{9}{tag-breath} is \mybox{boxgreen}{below \texttt{-21},} for which there is no physiological explanation.
We decide to remove this counterintuitive risk decrease by flattening~\inlinefig{10}{align-right} all scores below \texttt{-21}~(\autoref{fig:breath}\figpart{-B1}).
After this edit, we notice some fluctuations \mybox{boxblue}{on the right side} of the ``U-shape.''
Clinical knowledge suggests sepsis risk should only increase when \inlinefig{9}{tag-breath} increases for rates which are already above normal.
To fix the counterintuitive pattern in the model, we make the risk scores monotonically increasing~\inlinefig{9}{increasing} for bins between \texttt{7} and \texttt{40}~(\autoref{fig:breath}\figpart{-B2}).

\setlength{\columnsep}{6pt}%
\setlength{\intextsep}{0pt}%
\begin{wrapfigure}{R}{0.25\textwidth}
  \vspace{0pt}
  \centering
  \includegraphics[width=0.25\textwidth]{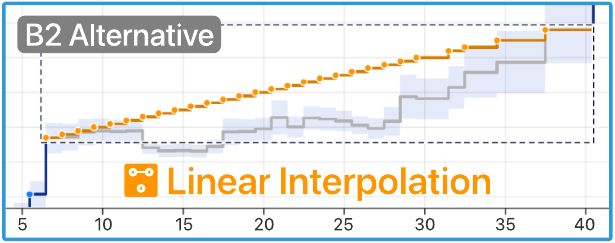}
\end{wrapfigure}
An alternative edit is to linearly interpolate~\inlinefig{9}{interpolate} the scores of bins from \texttt{7} to \texttt{40} (shown on the right).
However, we prefer the former edit, because linear interpolation~\inlinefig{9}{interpolate} would break the plateau of predicted risk when \inlinefig{9}{tag-breath} is between \texttt{8} and \texttt{28}, which are values that are commonly associated with children suffering from mild to moderate obstructive lung pathologies such as bronchiolitis and asthma, neither of which are likely to require treatment for suspected sepsis.
Removing this pattern might obscure a meaningful signal---there are many non-sepsis related reasons for moderately elevated respiratory rate.
Compared to the linear interpolation tool~\inlinefig{9}{interpolate}, the monotone increasing tool~\inlinefig{9}{increasing} is much less intrusive: it makes minimal changes to make the selected region monotone via isotonic regression.

\setlength{\belowcaptionskip}{-4.7pt}
\setlength{\abovecaptionskip}{7pt}
\begin{figure}[tb]
  \includegraphics[width=\linewidth]{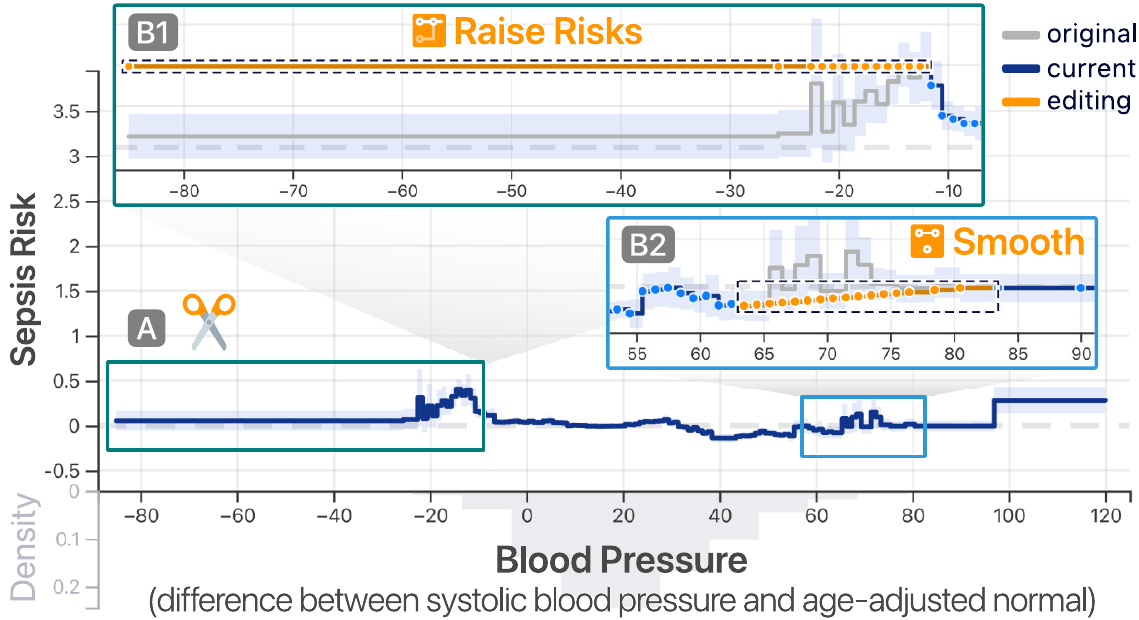}
  \Description{User interface for \tool{}}
  \caption{
    \inlinefig{9}{a3} Against physicians' expectations, a GAM predicts that patients with lower blood pressure have lower sepsis risk (left), and the risk abruptly increases at high blood pressure~(right).
    To create a safer model, \inlinefig{9}{b3-1}~we raise the risk scores~\inlinefig{10}{align-right}, and \inlinefig{9}{b3-1}~smooth out the sudden risk increase~\inlinefig{10}{interpolate}.
  }
  \label{fig:blood}
\end{figure}
\setlength{\belowcaptionskip}{0pt}
\setlength{\abovecaptionskip}{12pt}

\subsubsection{Editing the systolic blood pressure feature.}
The feature \inlinefig{9}{tag-pressure} measures the \textit{difference} between the systolic blood pressure in millimeters of mercury and its age-adjusted normal.
The \canvasview{}~(\autoref{fig:blood}\figpart{A}) shows that the model predicts patients with \inlinefig{9}{tag-pressure} from \texttt{-25} to \texttt{-10} to have a significantly higher risk of sepsis.
Interestingly, the predicted risk score decreases when \inlinefig{9}{tag-pressure} \mybox{boxgreen}{decreases after peaking at \texttt{-15}.}
The \canvasview{} shows only 19 patients out of 5000 patients with \inlinefig{9}{tag-pressure} below \texttt{-20}, and 118 patients with \inlinefig{9}{tag-pressure} from \texttt{-20} to \texttt{-10}.
Clinical knowledge suggests that when blood pressure readings move away from the typical range, both the odds of having a measurement artifact and the risk of sepsis increase.
To create a safer model, we select all the bins below \texttt{-15} and align~\inlinefig{10}{align-right} their risk score to the right~(\autoref{fig:blood}\figpart{-B1}).
Although by doing so, we raise the predicted risk score of all bins below \texttt{-15} to \texttt{0.38}, this is a conservative edit as we do not further increase the risk when \inlinefig{9}{tag-pressure} decreases after \texttt{-15}.
Here \inlinefig{9}{tag-pressure} below \texttt{-20} is most likely an error, and this edit might increase false-positive predictions on incorrect inputs.
However, the physician prefers this model to predict data errors and outliers as high risk, because it is safer to have a high false-positive rate than to have a high false-negative rate when predicting sepsis risk.
When editing healthcare models, physicians often consider the tradeoff between false-positive and false-negative rates, and the sweet spot for the tradeoff varies for different healthcare models~(see \autoref{sec:discussion:guideline} for more discussion).

The risk score of sepsis fluctuates when systolic \inlinefig{9}{tag-pressure} is \mybox{boxblue}{around \texttt{60}\textnormal{--}\texttt{80}}.
There is no physiological explanation for this fluctuation, so we smooth it out by linearly interpolating~\inlinefig{10}{interpolate} these scores.
Interestingly, there is a sudden increase in the predicted risk score when \inlinefig{9}{tag-pressure} is higher than 95, where these inputs are most likely errors.
Therefore, we decide not to edit this increase because it is safer to have a high false-positive rate than to have a high false-negative rate on a sepsis risk prediction model.
\subsection{Repairing Pneumonia Risk Prediction}
\label{sec:case-study-1}

KDD researchers~\cite{caruanaIntelligibleModelsHealthCare2015} have identified problematic patterns in pneumonia risk prediction models and raised the possibility to fix these patterns via model editing.
With \tool{}, we operationalize this possibility by editing the same model~\cite{caruanaIntelligibleModelsHealthCare2015} with a physician in our research team.
This GAM is trained to predict a patient's risk of dying from pneumonia.
The dataset includes 14,199 pneumonia patients; it has 46 features: 19 are continuous and 27 are categorical.
The outcome variable is binary: \texttt{1} if the patient died of pneumonia and \texttt{0} if they survived.
The AUC score on the test set (30\% of data) is \texttt{0.853}.
One ML researcher in our team loads \tool{} in their browser with 5,000 random training samples; they share their computer screen with a physician and 2 other researchers in the team via video-conferencing software.
All edits are made by the ML researcher after discussing with all people in the call.

\setlength{\belowcaptionskip}{0pt}
\setlength{\abovecaptionskip}{7pt}
\begin{figure}[tb]
  \includegraphics[width=\linewidth]{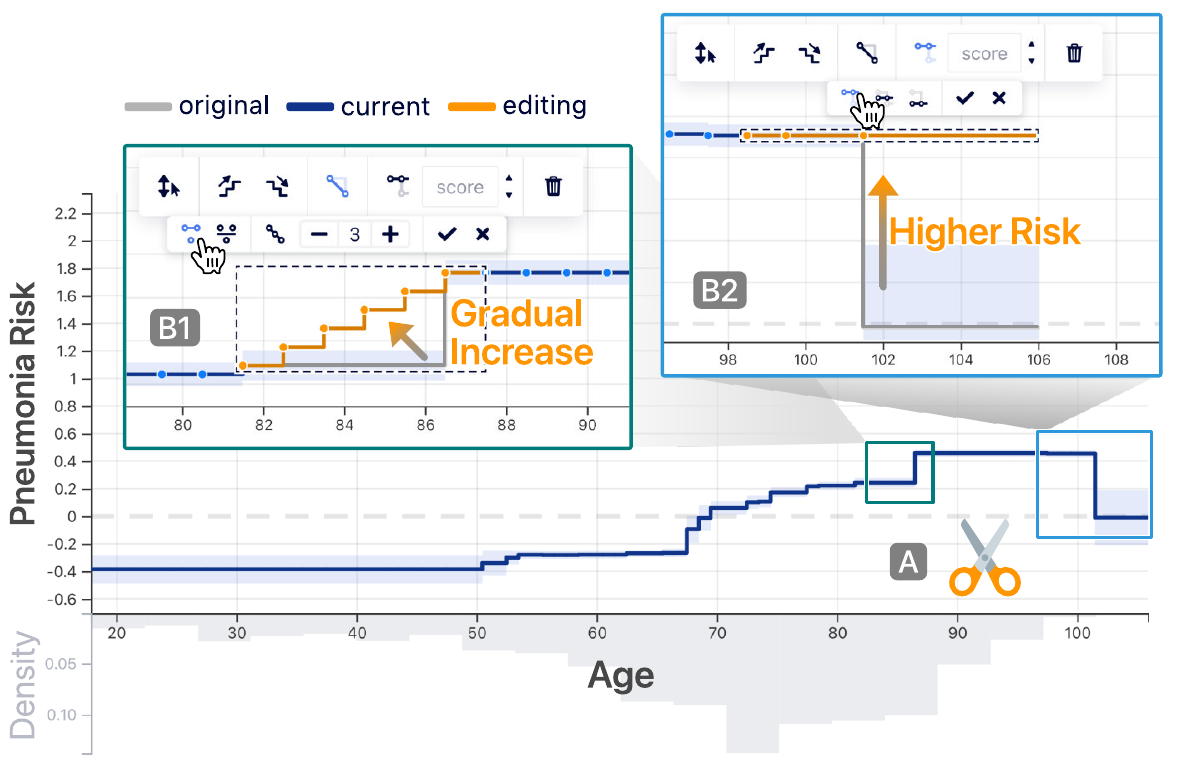}
  \Description{User interface for \tool{}, featuring four tightly integrated views:
    \canvasview{}, \metricview{}, \featureview{}, and \historyview{}
  }
  \caption{
    \inlinefig{9}{a3} Contrary to physicians' knowledge, a GAM predicts an abrupt increase of risk from age \texttt{86} to \texttt{87} (left), and that patients above \texttt{100} years old have lower pneumonia risk than patients \texttt{20} years younger (right).
    \inlinefig{9}{b3-1} With the interpolation tool \inlinefig{10}{interpolate}, we smooth out the abrupt increase of risk.
    \inlinefig{9}{b3-2} We use the align tool \inlinefig{10}{align} to raise the risk score for older patients.
  }
  \label{fig:scenario1}
\end{figure}
\setlength{\belowcaptionskip}{0pt}
\setlength{\abovecaptionskip}{12pt}

\subsubsection{Editing the age feature.}
After loading \tool{}, the \canvasview{}~(\autoref{fig:scenario1}\figpart{A}) first displays \inlinefig{9}{tag-age}, which has the highest importance score.
The x-axis ranges from \texttt{18} to \texttt{106} years old.
The y-axis encodes the predicted risk score (log odds) of dying from pneumonia.
It ranges from a score of \texttt{-0.4} for patients in their 20s to \texttt{0.5} for patients in their 90s.
The model predicts that younger patients have a lower risk than older patients.
However, the risk suddenly plunges when \mybox{boxblue}{patients pass \texttt{100}}---leading to a similar risk score as if the patient is \texttt{30} years younger!
It might be due to outliers in this \inlinefig{9}{tag-age} range, especially as this range has a small sample size, or patients who live this long might have ``good genes'' to recover from pneumonia.
To identify the true impact of \inlinefig{9}{tag-age} on pneumonia risk, additional causal experiments and analysis are needed.
Without robust evidence that people over 100 are truly at lower risk, physicians fear that they would be injuring patients by depriving needy older people of care, and violating their primary obligation to \textit{do no harm}.
Therefore, physicians would like to fix this pattern.
We apply a conservative remedy by setting~\inlinefig{10}{align} the risk of older patients to be equal to that of those slightly younger~(\autoref{fig:scenario1}\figpart{-B2}). \looseness=-1

\setlength{\columnsep}{6pt}%
\setlength{\intextsep}{0pt}%
\begin{wrapfigure}{R}{0.2\textwidth}
  \vspace{0pt}
  \centering
  \includegraphics[width=0.2\textwidth]{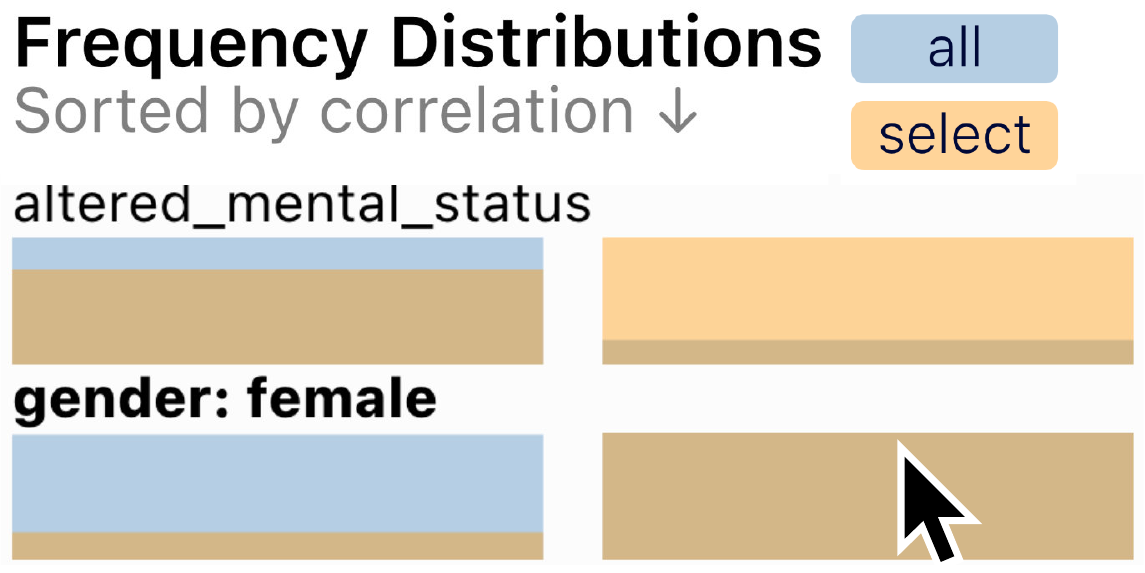}
\end{wrapfigure}
From the \metricview{}, we notice a drop of accuracy of 0.0004 in the \textit{Global Scope}, and the confusion matrix in the \textit{Selected Scope} shows that this edit causes the model to misclassify two negative cases as positives out of 28 patients who would be affected by the edit.
To learn more about these patients, we observe the \featureview{}, which shows that \inlinefig{9}{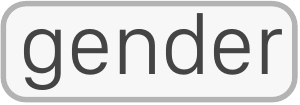} is the second most correlated categorical feature with the selected \inlinefig{9}{tag-age} range (shown on the right).
It means patients who are affected by this edit are disproportionally female---it makes sense because on average women live longer than men.
Seeing the correlated features helps us be aware of potential fairness issues during model editing.

Besides the problematic drop of risk for older patients, the risk suddenly rises \mybox{boxgreen}{around \texttt{86} years old}~(\autoref{fig:scenario1}\figpart{A}).
After converting the risk score from log-odds to probability, the predicted likelihood of dying from pneumonia increases by $4.89\%$ when the \inlinefig{9}{tag-age} goes from \texttt{86} to \texttt{87}.
This model behavior can cause \texttt{81}--\texttt{86} year-old patients to miss the care they need.
To create a safer model, we apply the linear interpolation tool~\inlinefig{10}{interpolate} in the region from \inlinefig{9}{tag-age} \texttt{81} to \texttt{87} to smooth out the sudden increase of pneumonia risk~(\autoref{fig:scenario1}\figpart{-B1}).

\setlength{\belowcaptionskip}{-5pt}
\setlength{\abovecaptionskip}{7pt}
\begin{figure}[tb]
  \includegraphics[width=\linewidth]{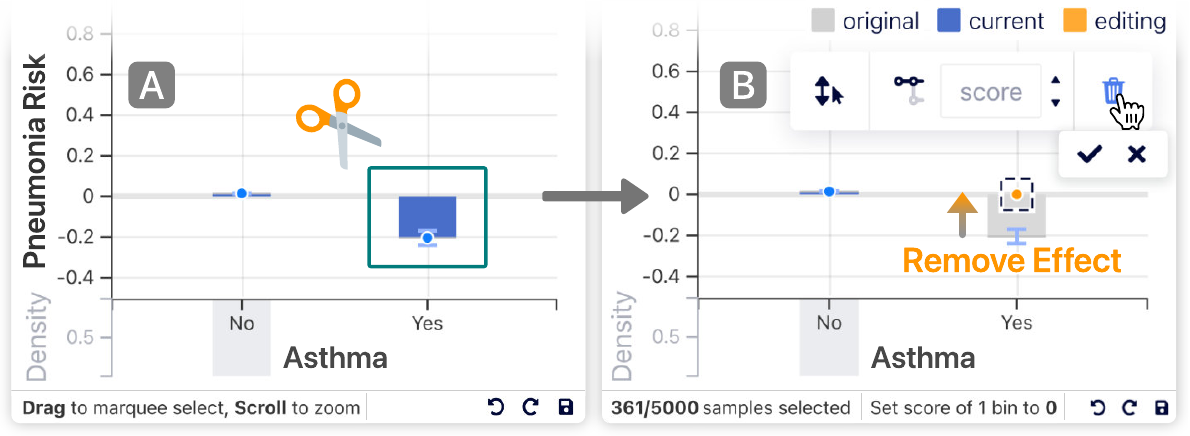}
  \Description{User interface for \tool{}, featuring four tightly integrated views:
    \canvasview{}, \metricview{}, \featureview{}, and \historyview{}
  }
  \caption{
    \inlinefig{9}{a3} A GAM predicts having asthma lowers the risk of dying from pneumonia.
    \inlinefig{9}{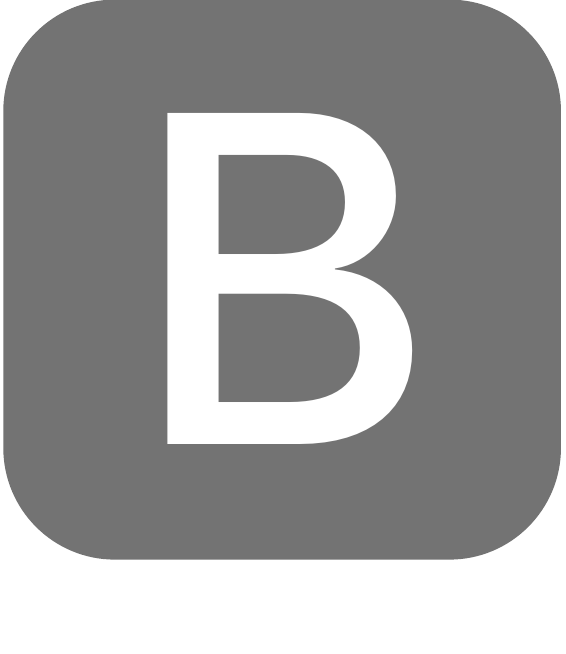} We address this problematic pattern by removing the predictive effect of having asthma~\inlinefig{10}{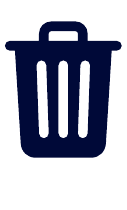}.
  }
  \label{fig:scenario2}
\end{figure}
\setlength{\belowcaptionskip}{0pt}
\setlength{\abovecaptionskip}{12pt}

\subsubsection{Editing the asthma feature.}
The \canvasview{}~(\autoref{fig:scenario2}\figpart{A}) of the binary feature \inlinefig{9}{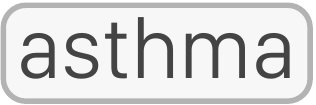} suggests the model predicts \mybox{boxgreen}{asthmatic patients} to have a lower risk of pneumonia than non-asthmatic patients.
It could be because pneumonia patients with a history of asthma are likely to receive care earlier and receive more intensive care.
However, if we use this model to make hospital admission decisions, this pattern might cause asthmatic patients to miss necessary care.
Therefore, we remove~\inlinefig{10}{delete} the predictive effect of having \inlinefig{9}{tag-asthma}~(\autoref{fig:scenario2}\figpart{B})---the new model would predict asthmatic patients to have an average risk.
This is a conservative edit as one might argue that asthmatic patients should have higher risk of pneumonia.
Our edit is a step in the right direction, but further experiments are needed to see if we need further adjustments.
\section{Impacts beyond healthcare}
\label{sec:user-study}
\textbf{Evaluation with data scientists.}
We conducted a user study to further evaluate the usability and usefulness of \tool{}, and also to investigate how data scientists would use our tool in practice.
In the study, we chose a loan default prediction model in a lending scenario, because there is no specialized knowledge needed to interpret and potentially edit this model.
The authors' Institutional Review Board~(IRB) has approved this study.

\subsection{Study Design}
\label{sec:study-design}

\textbf{Participants.}
The target users of \tool{} are ML practitioners and domain experts who are familiar with GAM models.
Therefore, we recruited 7 data scientists~(P1--P7) for this study by posting advertisements\footnote{Participant recruitment: \link{https://github.com/interpretml/interpret/issues/283}} on the online issue board of a popular GAM Library~\cite{noriInterpretMLUnifiedFramework2019}.
The participation was voluntary, and we did not collect any personal information.
All participants have developed GAMs for work: three participants use GAMs multiple times a week~(P1, P5, P6), three use them a few times a month~(P2, P3, P4), and one uses them about once a month~(P7).
Four participants work in finance~(P1, P2, P3, P7), two work in healthcare~(P4, P5), and one works in media~(P6).
Each study lasted about 1 hour, and we compensated each participant with a \$25 Amazon Gift card.

\textbf{Procedure.}
We conducted the study with participants one-on-one through video-conferencing software.
With permission from all participants, we recorded the video conference for subsequent analysis.
After signing a consent form and a background questionnaire (e.g., familiarity with GAMs), each participant was given an 8-minute tutorial about \tool{}.
Participants then were pointed to a website consisting of \tool{} with a model trained on the LendingClub dataset~\cite{LendingClubOnline2018} to predict if a loan applicant can pay off the loan: the outcome variable is \texttt{1} if they can and \texttt{0} otherwise (see \autoref{appendix:user-study} for details).
Participants were given a list of recommended tasks to look for surprising patterns, edit 3 continuous features and 2 categorical features with different editing tools, experiment with different views, and freely explore the tool.
Participants were told that the list was a guide to help them try out all features in the tool, and they were encouraged to freely edit the model as seen fit.
Participants were asked to think aloud and share their computer screens with us.
Each session ended with a usability survey and a semi-structured interview that asked participants about their experience of using \tool{} and if this tool could fit their workflow and help them improve models in practice.

\subsection{Benefits to Data Scientists}
\label{sec:study-results}

Below we summarize key findings from our observations and participants' qualitative feedback.

\subsubsection{Meet the pressing needs for model editing}
\label{sec:study-needs}

Through analyzing interviews and participants' verbalization of thoughts during the exploration task,
we find there are critical needs for model editing in practice, and ML practitioners have already been editing their models with different methods.
All participants have observed counterintuitive patterns when developing models in their work.
For example, P6 recalled their GAM model, \myquote{Some weights are negative, and I know by definition this cannot happen because [... of the nature of that feature].}
P7 commented \myquote{[Strange patterns] happen a lot, mostly the direction of a certain variable. We expect the score to be increasing; however, the model shows something opposite.}

Many participants were required to fix these strange patterns.
P3 and P7 needed to remove counterintuitive patterns because of the \textit{Adverse Action Notice Requirement}, a policy requiring lenders to provide explanations to loan applicants.
If there are strange patterns, the model explanations sometimes will not make sense to loan applicants.
P7 explained, \myquote{Basically you want to make the model easier to explain in adverse action calls.}
Adverse action calls refer to situations when applicants dial in and demand real-time model explanations.
On the other hand, P5 and P6 needed to edit their models on some well-understood features to align model behaviors with the expectations of knowledgeable stakeholders---physicians and business partners, respectively.
In addition, P1 edited their models because they found enforcing monotonicity and removing small bumps had improved model accuracy in deployment.

\textbf{Improve and unify current editing approaches.}
Most participants reported using feature engineering to fix counterintuitive patterns in their own day-to-day work.
For example, after discussing with domain experts, P5 removed features where they thought the shape functions were wrong or did not make sense.
In P7's case, a legal compliance team would decide which features to include and exclude after inspecting the model behaviors.
P2 trained multiple models with different hyper-parameters and then chose models that not only had high accuracy but also learned expected trends.
P1 had set up a sophisticated post-processing pipeline that would automatically smooth out shape functions, enforce monotonicity, and remove predictive effects on missing values.
With interactivity and flexible tools, \tool{} provided participants with direct control of their model behaviors and unify current editing approaches. \looseness=-1

\subsubsection{Usable and useful}
\label{sec:study-usability}

\begin{figure}[tb]
  \includegraphics[width=0.92\linewidth]{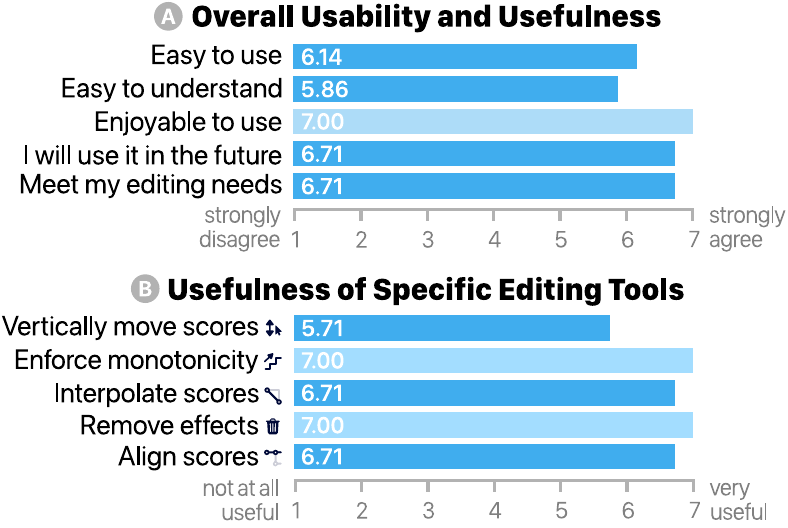}
  \Description{User interface for \tool{}}
  \caption{
    Average ratings from 7 participants for \tool{}'s usability and usefulness.
    (A)
    All participants enjoyed using the tool; they found it highly usable and it meets their editing needs.
    (B) All features, especially enforcing monotonicity and removing effects, were rated favorably.
  }
  \label{fig:study}
\end{figure}

The study survey included a series of 7-point Likert-scale questions regarding the usability and usefulness of \tool{}~(\autoref{fig:study}\figpart{A}).
The results suggest that the tool is easy to use~(average 6.14), easy to understand~(average 5.86), and especially enjoyable to use~(average 7.00---all participants gave the highest rating).
Most participants would like to use \tool{} in their work to edit models.
For example, P6 commented \myquote{I have the dire hope that it will be a groundbreaking experience. [...] I strongly believe that this interactive model editing will please a lot of stakeholders, and increases trust and acceptance.}

\setlength{\belowcaptionskip}{-8pt}
\setlength{\abovecaptionskip}{4pt}
\setlength{\belowcaptionskip}{0pt}
\setlength{\abovecaptionskip}{12pt}

\textbf{Versatile editing tools.}
We asked participants to rate specific editing tools in \tool{}~(\autoref{fig:study}\figpart{B}).
All tools were rated favorably, and participants particularly liked the monotonicity tool~\inlinefig{10}{increasing} and deletion tool~\inlinefig{10}{delete} (both received the highest rating from all participants).
Monotonicity constraints are common across different domains, which might explain the high interest in the monotonicity tool.
In particular, P4 appreciated that the monotonicity tool supported regional monotonicity: P4 gave an example from his work where the relationship between the \inlinefig{9}{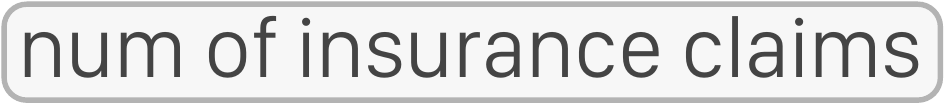} and people's \inlinefig{9}{tag-age} was expected to form a ``U-shape'' (kids and seniors tend to have more insurance claims), and he would like to use our tool to enforce monotonicity with different directions~\inlinefig{10}{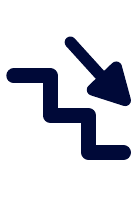}~\inlinefig{10}{increasing} on the two ends of the shape function.
Unlike the monotonicity tool, the deletion tool~\inlinefig{10}{delete} had a much simpler functionality, and yet it was participants' favorite.
P7 liked the deletion tool because it was useful to edit categorical features, \myquote{For missing values and neutral values [in categorical features], we don't want to reward them, and we don't want to punish them, so we usually just neutralize them [with the deletion tool].}
Participants' overwhelmingly positive feedback provides evidence that \tool{} is easy to use, and it can help practitioners improve their ML models through model editing.

\subsubsection{Fit into model development workflows}
\label{sec:study-workflow}

Interviews with participants highlight that \tool{} fits into data scientists' workflows.
Five participants used Jupyter notebooks to develop ML models, and they all appreciated that they could use \tool{} directly in their notebooks.
Many participants found the ``git commit'' style of editing history in the \historyview{}~(\autoref{sec:experience:history}) familiar and useful.
When P6 wrote edit commit messages, they followed their company's git commit style to include their name and commit type at the end of the message.
In addition, P3 found the editing history and auto-generated messages helpful for their company's model auditing process, \myquote{I especially like the history panel where all the edits are tracked. You can technically use it as a reference when writing your model documentation [for auditors to review].}

\textbf{A platform for collaboration.}
Interestingly, many participants commented that besides model editing, \tool{} would be a helpful tool to communicate and collaborate with different stakeholders.
For example, P5's work involved collaborating with physicians to interpret models, and they thought our tool would be a tangible tool to promote discussion about models: \myquote{This work is very important because it lets people discuss about it [model behaviors].}
P1 had been building dashboards to explain models to their marketing teams, and they would like to use \tool{} to facilitate the communication.
Similarly, P6 told us they would use our tool to communicate model insights to their stakeholders, including business partners, UX designers, and the sales team.

\subsubsection{Diverse ways to use \tool{}}
\label{sec:study-diverse}

Even with a relatively small sample size of 7 participants, we observed a wide spectrum of views regarding \textit{when} and \textit{how} to edit models.
For example, P2 was more conservative about interactive model editing; they felt it was more ``objective'' to retrain the model until it learned expected patterns rather than manually modifying the model.
P3 thought \tool{} would be useful to enforce monotonicity and fix obvious errors, but they were more cautious and worried about irresponsible edits: \myquote{Anyone behind the model can just add whatever relationship they want, rather than keep the model learn empirically whatever is in the data. I mean, it [the tool] is good, but you need to be diligent and make sure whatever changes you made make sense and are justifiable.}
On the other side of the spectrum, P5 and P6 found model editing with \tool{} very natural as they had already been iterating on models with domain experts.

\setlength{\columnsep}{5pt}%
\setlength{\intextsep}{0pt}%
\begin{wrapfigure}{R}{0.21\textwidth}
  \vspace{0pt}
  \includegraphics[width=0.21\textwidth]{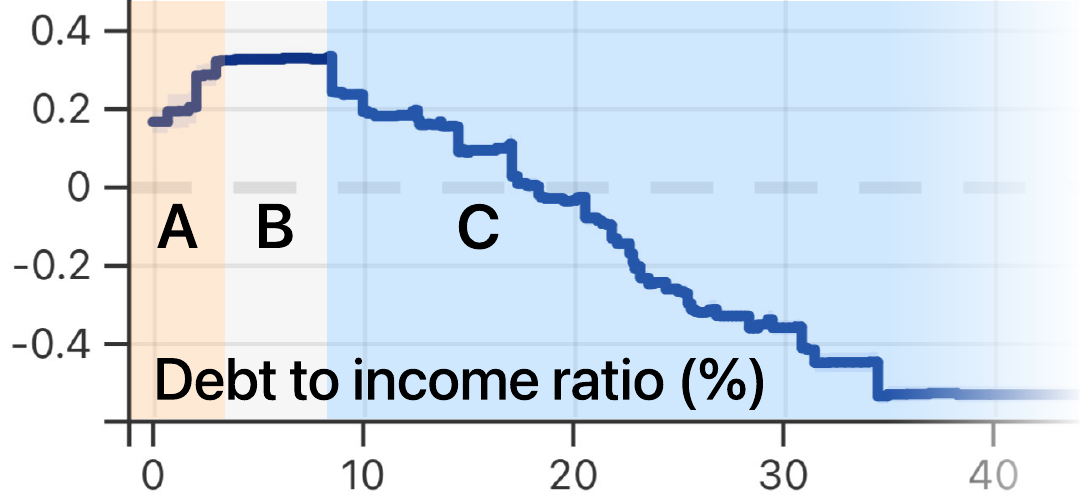}
  \vspace{-10pt}
\end{wrapfigure}
\textbf{Multiple approaches.}
In addition to \textit{whether} and \textit{when} people should edit models, participants had different views on \textit{how} to edit the model.
For example, in the model used in this user study, debt to income ratio (\inlinefig{9}{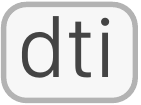}) is a continuous feature (shown on the right): the log odds score (y-axis) of an applicant paying off their loan first increases when \inlinefig{9}{tag-dti} (x-axis) increases from \texttt{0\%} to \texttt{3\%} (\orangelighthl{area A}); after a plateau (\graylighthl{area B}), the score then decreases when \inlinefig{9}{tag-dti} increases from \texttt{8\%} to \texttt{40\%} (\bluelighthl{area C}).
One suggested task is to increase the score for low \inlinefig{9}{tag-dti} in \orangelighthl{area A}.
Five participants (P1, P2, P3, P4, and P7) commented the trend in \orangelighthl{area A} made sense---applicants in this range are likely people who have no or little loan experience and thus less likely to pay off the loan in time.
Although the pattern made sense to P3 and P7, they agreed that one should fix it;
P3 and P7 raised the score by aligning~\inlinefig{10}{align-right} all scores in \orangelighthl{area A} to be the same as \graylighthl{area B}.
P3 explained: \myquote{[Although the pattern in area A makes sense,] we'll still try to make this relationship monotonic. For the relationship that I described, like somebody is less experienced with the credit and other stuff, there are other variables that will factor in, like the number of accounts open.}
P7 made the same edit but with a different reason: \myquote{We do not want a model to punish people with no debt.}
In contrast to P3 and P7, P4 said they were uncomfortable with raising the scores in \orangelighthl{area A}, and they would need to talk to finance experts if they were editing this model in practice.
P1 also decided to keep the trend in \orangelighthl{area A}.
Additionally, P1 applied the interpolation tool~\inlinefig{10}{interpolate} to smooth the score increase in \orangelighthl{area A} and decrease in \bluelighthl{area C}, because P1 believed small bumps in \orangelighthl{area A} and \bluelighthl{area C} are due to overfitting.
Participants' diverse views on \textit{whether}, \textit{when}, and \textit{how} to edit models highlight that users with different backgrounds may use \tool{} differently in practice.\looseness=-1

\headermargintop{}
\section{Discussion and Future Work}
\label{sec:discussion}

Reflecting on our iterative design of \tool{} with diverse stakeholders, model editing experiences with physicians, and an evaluation with data scientists in various domains, we distill lessons and future directions for model editing and interpretability research.

\textbf{Promote accountable edits \& develop guidelines.}\xlabel[]{sec:discussion:guideline}
Our user study shows model editing via feature engineering and parameter tuning is already common practice in data scientists' workflow~(\autoref{sec:study-needs}).
As the first interactive model editing tool, \tool{} lowers the barrier to modifying model behaviors to reflect users' domain knowledge and values.
We find different users could have distinct views on \textit{whether}, \textit{when}, and \textit{how} to edit models~(\autoref{sec:study-diverse}).
Some users might raise concerns that \tool{} makes model editing too easy, and that irresponsible edits could potentially cause harm~(e.g., P3 in \autoref{sec:study-diverse}).
Guarding against harmful edits is our top priority---we provide users with continuous feedback~(\autoref{sec:experience:method}), as well as transparent and reversible edits~(\autoref{sec:experience:responsible}).
However, they do not guarantee to prevent users from overfitting the model, injecting harmful bias, or maliciously manipulating model predictions.
This potential vulnerability warrants further study on how to audit and regulate model editing.

To help ``model editors'' modify ML models responsibly, we see a pressing need of \textit{guidelines} that unify best practices in model editing.
However, model editing is complex---\textit{whether}, \textit{when}, and \textit{how} to edit a model depends on many factors, including the data, model's behaviors, and end-tasks in a sociotechnical context.
Take our sepsis risk prediction model as an example~(\autoref{sec:case-study-2}); we inform our edit decisions by considering treatment effects, the potential impact of edits, and physicians' values.
We make specific edits because physicians prefer false positives over false negatives when predicting sepsis risks---we will make different edits if false negatives are favored.
For example, in prostate cancer screenings, false positives are much riskier than false negatives~\cite{croswellPrinciplesCancerScreening2010}.
Therefore, we may prioritize lowering the predicted risk when fixing problematic patterns in a risk prediction model for prostate cancer.
Using \tool{} as a research instrument, we plan to develop editing guidelines by further research that engages with experts in diverse domains as well as people who would be impacted by edited models.

\vspace{2pt}
\textbf{Measure real-life impacts.}\xlabel[]{sec:discussion:impact}
\tool{} provides continuous feedback on model performance~(\autoref{sec:experience:responsible}).
Due to the additive nature of GAMs, global metrics---computed on all validation samples---are not very sensitive to edits that slightly change a few bins of a single feature.
An edit's effect is more significant when we measure the accuracy locally, such as in the \textit{Selected Scope} or the \textit{Slice Scope}.
The \metricview{}'s goal is to alert users of accidental edits that might demolish the model's predictive power or disproportionally affect a subgroup in the data.
However, \tool{}'s ultimate goal is to help users create \textit{safer} and \textit{more correct} models---accuracy on the train and test sets is a secondary metric.
To evaluate model editing, we need to measure edited models' performance for their intended use.
In high-stakes settings such as healthcare, editing would make a substantial impact if it changed a deployed model's prediction on one patient.
We plan to adapt the edited sepsis risk prediction model~(\autoref{sec:case-study-2}) in a large hospital and conduct a longitudinal study to monitor and investigate the model's performance.

\vspace{2pt}
\textbf{Enhance collaborative editing.}\xlabel[]{sec:discussion:collaboration}
When using \tool{} to edit healthcare models with physicians, we find the tool provides a unique \textit{collaborative experience} for ML researchers and domain experts to discuss, interpret, and improve models together.
Our user study echos this observation: (1) participants had been editing models through teaming with diverse stakeholders including domain experts, auditors, and marketing teams~(\autoref{sec:study-needs}); (2) participants appreciated \tool{} as a platform to facilitate ML communication with various stakeholders~(\autoref{sec:study-workflow}).
Therefore, we would like to further enhance the tool's affordance for collaborations.
We plan to explore interaction techniques that support multiple users to edit the same model simultaneously (e.g., Google Slides).
Also, we plan to enhance our Git-inspired editing history to support users to \textit{merge} multiple independent edit series onto one model---enabling collaborators to easily edit a model asynchronously.
\section{Conclusion}

In this work, we present \tool{}, an interactive visualization tool that empowers domain experts and data scientists to not only interpret ML models, but also align model behaviors with their knowledge and values.
This open-source tool runs in web browsers or computational notebooks, broadening people's access to responsible ML technologies.
We discuss lessons learned from two editing examples and an evaluation user study.
We hope our work helps emphasize the critical role of human agency in responsible ML research, and inspire future work in actionable ML interpretability.

\begin{acks}
The first two authors were interns at Microsoft Research.
We are grateful to Scott Lundberg, Steven Drucker, Adam Fourney, Saleema Amershi, Dean Carignan, Rob DeLine, Haekyu Park, the InterpretML team, and our study participants for their helpful feedback.
\end{acks}

\begin{spacing}{0.975}
\bibliographystyle{ACM-Reference-Format}
\bibliography{gam-changer}
\end{spacing}

\appendix
\clearpage

\setcounter{figure}{0}
\renewcommand{\thetable}{S\arabic{table}}
\renewcommand{\thefigure}{S\arabic{figure}}

\section{\tool{} User Interface}
For reproducibility, we discuss the design and implementation of \tool{} in detail.
The tool's interface is based on our early prototype~\cite{wangGAMChangerEditing2021}.
We use a GAM regression model trained on the public Iowa house price dataset~\cite{decockAmesIowaAlternative2011} to illustrate the interface.

\toggletrue{inheader}
\xlabel[\ddag A.1]{sec:canvasview}
\subsection{\canvasview{}}
\togglefalse{inheader}

\begin{figure*}[tb]
  \includegraphics[width=\linewidth]{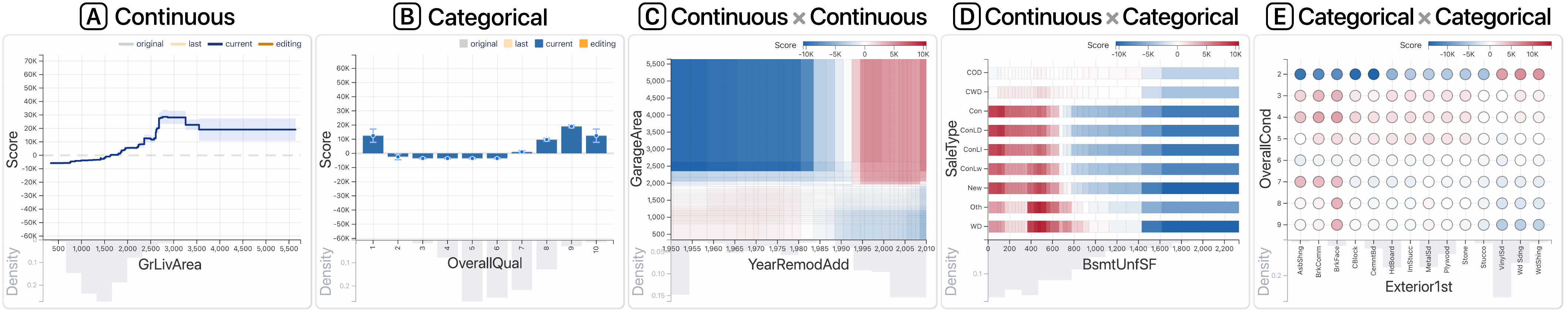}
  \Description{User interface for \tool{}}
  \caption{
    The \canvasview{} employs different designs to visualize shape functions on different feature types.
    We use \inlinefig{9}{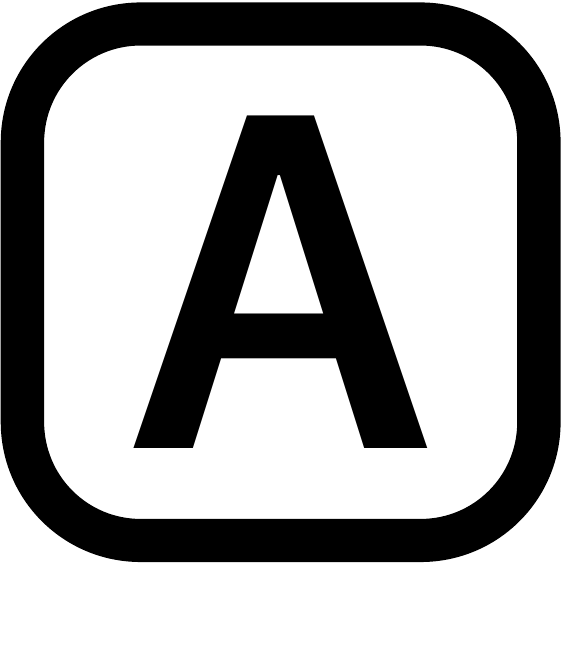} line charts for continuous variables, \inlinefig{9}{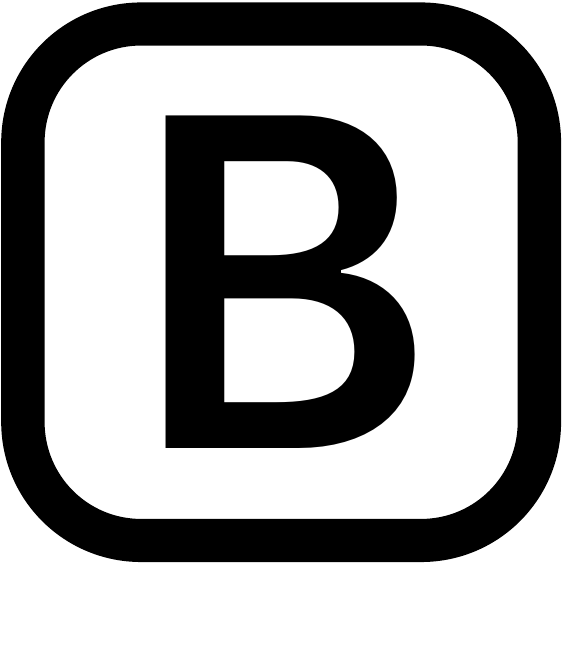} bar charts for categorical variables, \inlinefig{9}{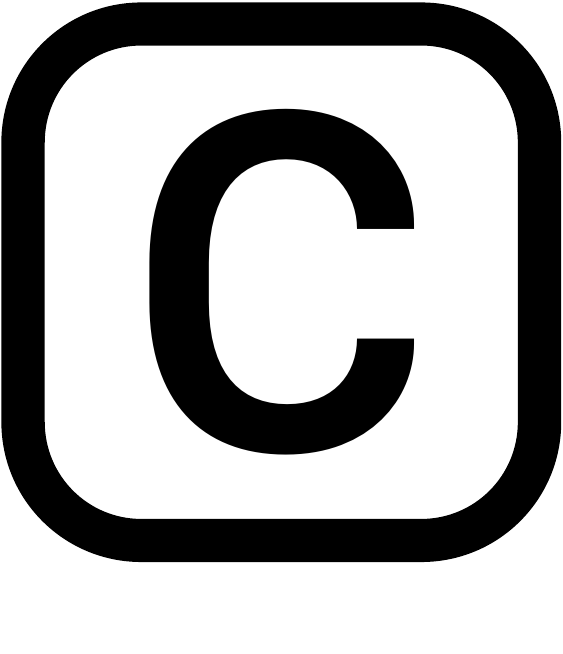} heatmaps for interaction effects of two continuous variables,
    \inlinefig{9}{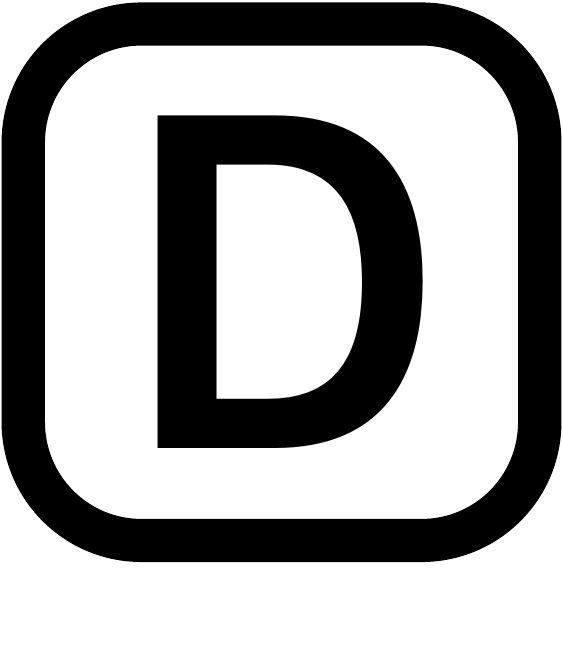} vertical bar charts for interaction effects between continuous and categorical variables,
    and \inlinefig{9}{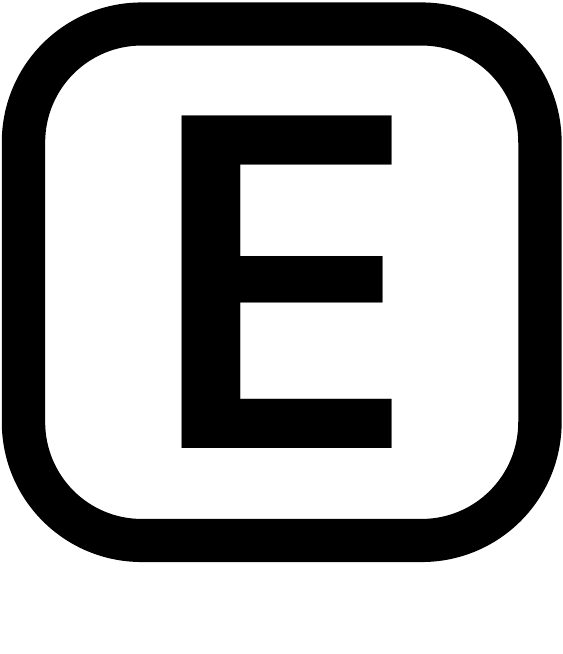} scatter plots for interaction effect of two categorical variables.
    For univariate features, the x-axis encodes the input feature $x_j$, and the y-axis represents the output of the shape function $f_j(x_j)$.
    We also use light-blue bands and error bars to represent the prediction confidence.
    For pair-wise interactions, the axes encode two features, and we use a diverging color scale to represent the contribution scores.
  }
  \label{fig:gallery}
\end{figure*}

\setlength{\columnsep}{8pt}%
\setlength{\intextsep}{0pt}%
\begin{wrapfigure}{R}{0.14\textwidth}
  \vspace{0pt}
  \centering
  \includegraphics[width=0.14\textwidth]{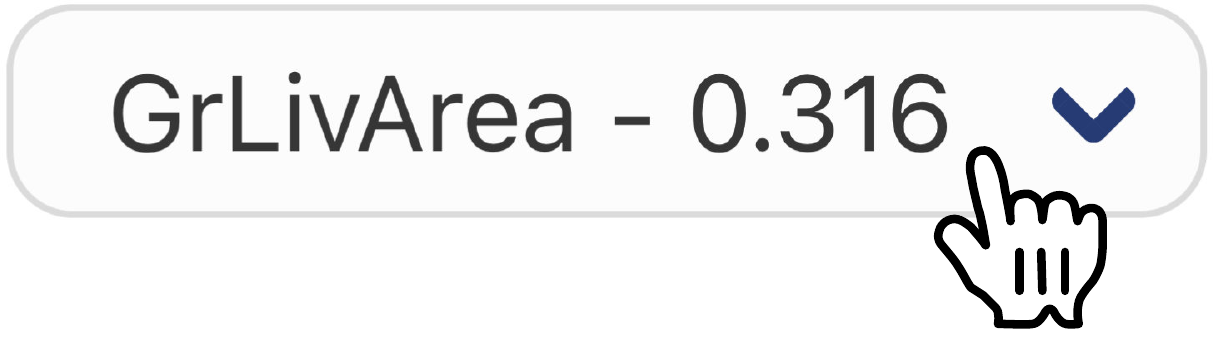}
\end{wrapfigure}
In the \canvasview{}~(\autoref{fig:teaser}\figpart{A}), users can inspect and direct manipulate shape functions.
As GAMs support continuous and categorical features, as well as their two-way interactions, we design unique visualization for each variable type, featuring line chart, bar chart, heatmaps, and scatter plots~(\autoref{fig:gallery}).
Users can use the feature selection drop-down to transition across features.
To begin, the \canvasview{} shows the feature with the highest importance score, computed as the weighted average of a feature's absolute contribution scores.
We re-center the contribution scores by adjusting the \tcolor{mygreen}{intercept constant $\beta_0$}~(\autoref{equation:gam}) such that the mean prediction for each feature has a zero score across the training data.
Thus, a positive score suggests the feature positively affects the prediction and vice versa.
Consider a GAM trained to predict house prices~(\autoref{fig:gallery}\figpart{A}), if the living area is larger than 2000 square feet, it increases the predicted house price, while areas lower than 2000 decrease the predicted value compared with average.
We highlight the $0$-baseline as a thick dashed line.

\begin{figure}[b]
  \includegraphics[width=0.9\linewidth]{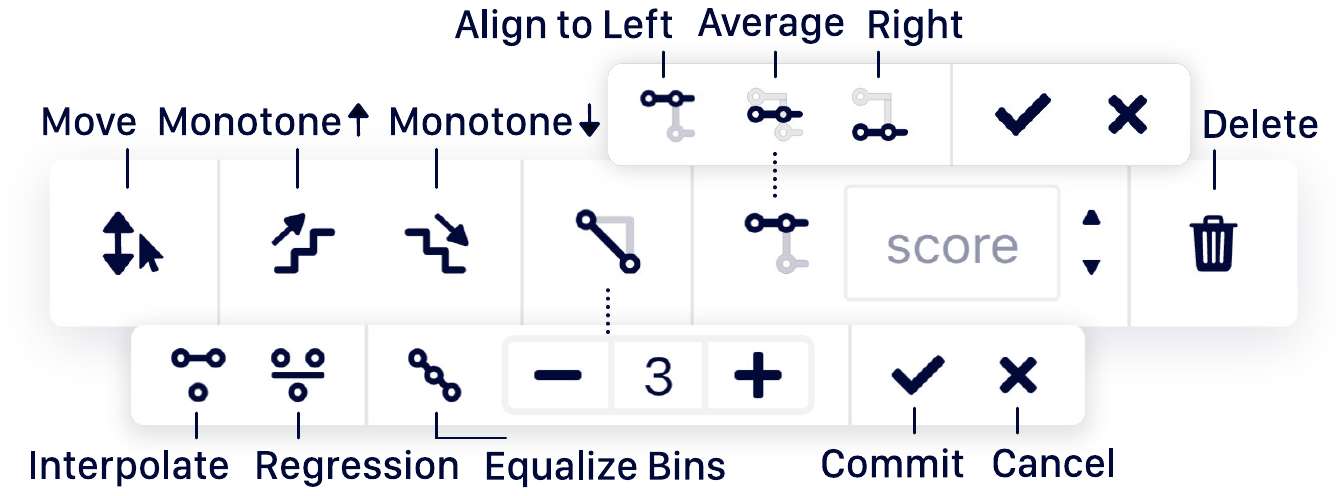}
  \Description{User interface for \tool{}}
  \caption{
    The \toolbar{} enables users to edit GAMs with a variety of editing tools.
    Users can use the move tool~\inlinefig{10}{move} to adjust the contribution scores of selected bins by dragging bins up and down.
    Users can apply the interpolate tool~\inlinefig{10}{interpolate} to linearly interpolate the scores of an interval of bins from the start to the end.
    Alternatively, users can interpolate scores with an arbitrary number of equal bins~\inlinefig{10}{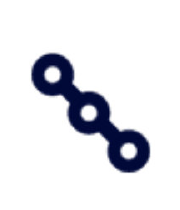}, or by fitting a linear regression~\inlinefig{10}{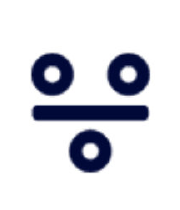}.
    With minimal changes, the monotonicity tool transforms the selected scores into a monotonically increasing function~\inlinefig{10}{increasing} or a monotonically decreasing function~\inlinefig{10}{decreasing}.
    With align tools, users can unify the selected scores as the score of the left bin~\inlinefig{10}{align}, the right bin~\inlinefig{10}{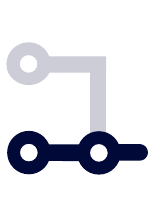}, or the average score weighted by the training sample counts~\inlinefig{10}{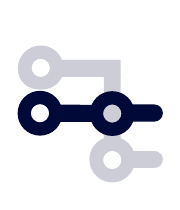}.
    Users can also use the delete tool~\inlinefig{10}{delete} to set all selected scores to \texttt{0}.
    }
  \label{fig:toolbar}
\end{figure}

\setlength{\columnsep}{6pt}%
\setlength{\intextsep}{0pt}%
\begin{wrapfigure}{R}{0.15\textwidth}
  \vspace{0pt}
  \centering
  \includegraphics[width=0.15\textwidth]{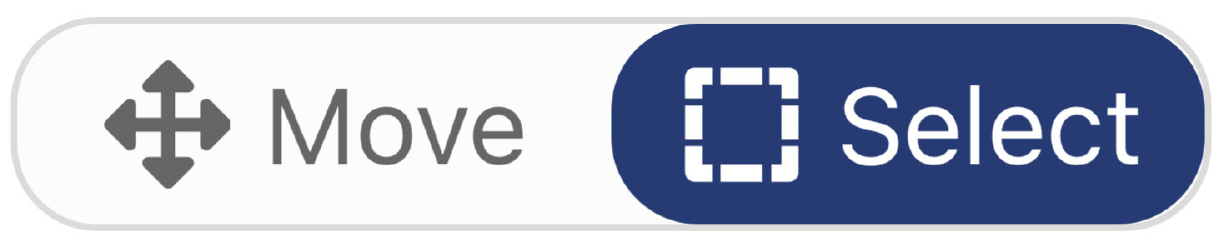}
\end{wrapfigure}
\xlabel[\ddag~A.1]{appendix:tools}
\textbf{Editing tools.}
In the \canvasview{}, users can switch between \textit{move mode} and \textit{select mode} by clicking the mode toggle button.
In the \textit{move mode}, users can use \textit{zoom-and-pan} to control their view portion and focus on analyzing an interesting region in the GAM visualization.
In the \textit{select mode}, users can use \textit{marquee selection} to pick a subset of bins (or bars for categorical features) to edit.
Once a region of the shape function is selected, the \toolbar{}~(\autoref{fig:toolbar}) appears.
In the bottom \statusbar{}, users can view the number of samples in the selected region and a description of their last edit.
Users can click the check icon \inlinefig{9}{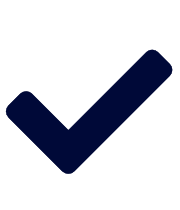} to ``commit''~(\autoref{sec:experience:history}) the change if they are satisfied with this edit, or click the cross icon \inlinefig{9}{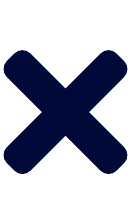} to discard the change.

\toggletrue{inheader}
\xlabel[\ddag~A.2]{sec:featureview}
\subsection{\featureview{}}
\togglefalse{inheader}

\setlength{\belowcaptionskip}{-1pt}
\setlength{\abovecaptionskip}{7pt}
\begin{figure}[b]
  \includegraphics[width=\linewidth]{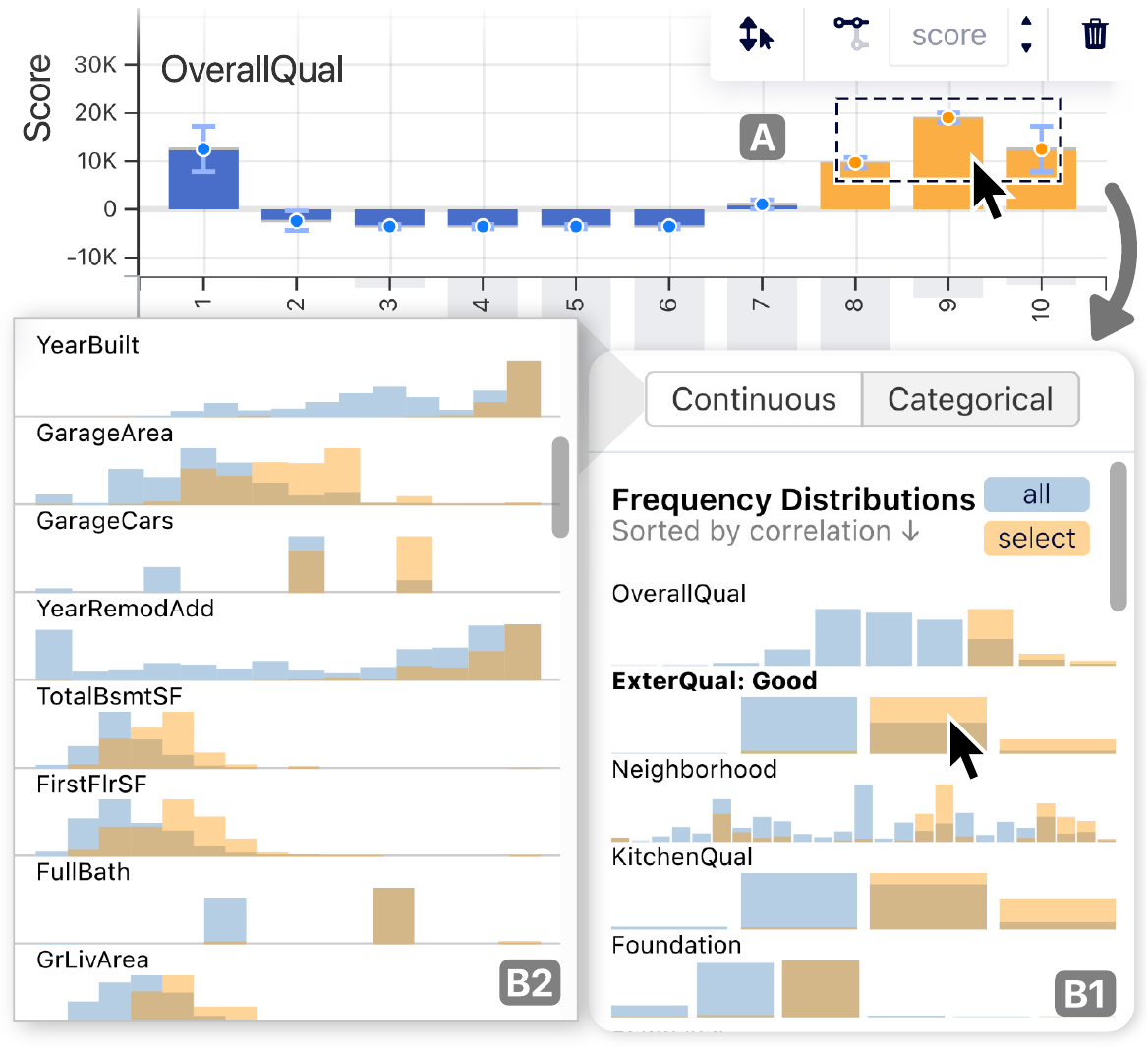}
  \Description{User interface for \tool{}, featuring four tightly integrated views:
    \canvasview{}, \metricview{}, \featureview{}, and \historyview{}
  }
  \caption{
    \inlinefig{9}{a3} On a GAM trained to predict house price, a user selects bins representing houses with high quality in the \canvasview{}.
    \inlinefig{9}{b3-1} For categorical variables, the \featureview{} shows that selected houses disproportionally have better exterior and kitchen quality and locate in certain neighborhoods.
    \inlinefig{9}{b3-2} For continuous variables, the year built and garage area are also highly correlated with the house quality.
  }
  \label{fig:correlation}
\end{figure}

\setlength{\belowcaptionskip}{0pt}
\setlength{\abovecaptionskip}{12pt}

The \featureview{}~(\autoref{fig:teaser}\figpart{-B2}, \autoref{fig:correlation}) helps users gain an overview of correlated features as well as their distributions and elucidate potential editing effect disparities.
We develop \textit{linking+reordering}---a novel technique to identify correlated features.
Once a user selects an interval of the shape function in the \canvasview{}~(\autoref{fig:correlation}\figpart{-A}), we look up affected samples and their associated bins across all features.
For each feature, we compare the bin count frequency in \bluelighthl{all training data} and the frequency in the \orangelighthl{\strut selected samples} by computing the $\ell_2$ distance between these two frequency vectors.
Then, we plot two frequency distributions in an overlaid histogram for each feature, and sort all histograms in descending order of the distance scores~(\autoref{fig:correlation}\figpart{-B}).
The intuition is that if two features $x_1$ and $x_2$ are independent, then samples selected from an interval in $x_1$ should have a \orangelighthl{distribution} similar to the \bluelighthl{training data distribution} in $x_2$, and vice versa.
Therefore, correlated features will be on top of the sorted histogram list.
Our \textit{linking+reordering} technique allows users to interactively and quickly identify local correlations across features, even between continuous and categorical features.
By observing correlated features, users can identify potential disparities in editing impacts.
For example, editing high-quality houses would disproportionately affect newer houses~(\autoref{fig:correlation}).

\toggletrue{inheader}
\xlabel[\ddag~A.3]{sec:metricview}
\subsection{\metricview{}}
\togglefalse{inheader}

\setlength{\columnsep}{8pt}%
\setlength{\intextsep}{0pt}%
\begin{wrapfigure}{R}{0.19\textwidth}
  \vspace{0pt}
  \centering
  \includegraphics[width=0.19\textwidth]{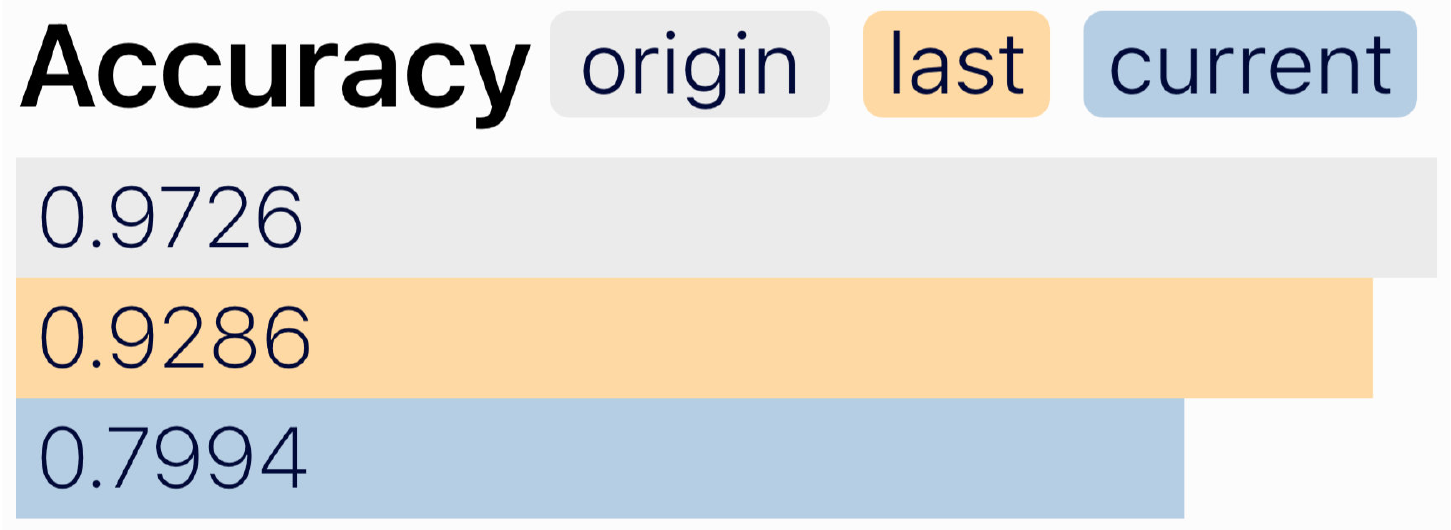}
\end{wrapfigure}
The \metricview{}~(\autoref{fig:teaser}\figpart{-B1}) provides real-time and continuous feedback on the model performance.
For a binary classifier, we present a confusion matrix and use bar plots to encode the model's accuracy, balanced accuracy, and the Area Under the Curve (AUC).
For a regressor, we report root mean squared error, mean absolute error, and mean absolute percentage error.
We use the same color codes of shape functions in the \canvasview{} to describe the performance of the original model, the model from the last edit, and the current model.

\setlength{\columnsep}{8pt}%
\setlength{\intextsep}{0pt}%
\begin{wrapfigure}{R}{0.19\textwidth}
  \vspace{3pt}
  \centering
  \includegraphics[width=0.19\textwidth]{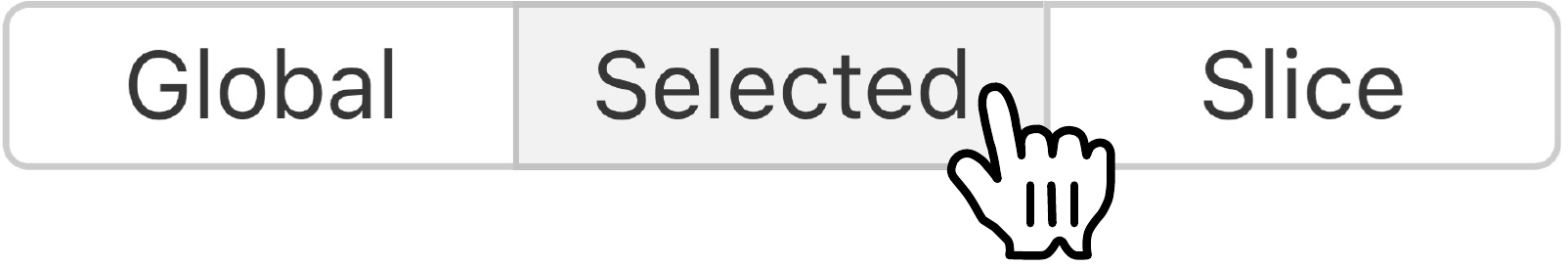}
\end{wrapfigure}
Besides monitoring global metrics that are computed on all validation samples, users can choose a subset of validation samples to compute the metrics by switching the metric scope.
For example, with the \textit{Selected Scope}, the \metricview{} only computes model metrics on samples that are in the currently selected region.
With the \textit{Slice Scope}, users can choose a data slice by selecting a level of a categorical variable, e.g., the \inlinefig{9}{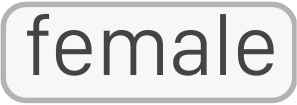} level of the \inlinefig{9}{tag-gender} variable.
Then, performance metrics in the \metricview{} are computed on the validation samples that belong to the selected subgroup.

\toggletrue{inheader}
\xlabel[\ddag~A.4]{sec:historyview}
\subsection{\historyview{}}
\togglefalse{inheader}

\setlength{\columnsep}{5pt}%
\setlength{\intextsep}{0pt}%
\begin{wrapfigure}{R}{0.07\textwidth}
  \vspace{-3pt}
  \centering
  \includegraphics[width=0.07\textwidth]{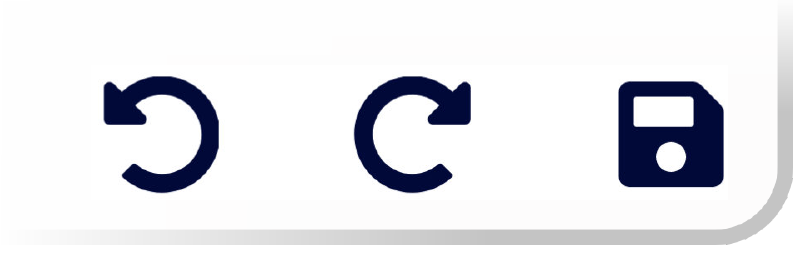}
\end{wrapfigure}
In \tool{}, users can easily undo and redo their edits by clicking the buttons in the bottom \statusbar{} (shown on the right) or using keyboard shortcuts.
Reversible actions promote accountable model editing, as users can easily fix their mistakes.

\setlength{\columnsep}{9pt}%
\setlength{\intextsep}{0pt}%
\begin{wrapfigure}{R}{0.2\textwidth}
  \vspace{0pt}
  \centering
  \includegraphics[width=0.2\textwidth]{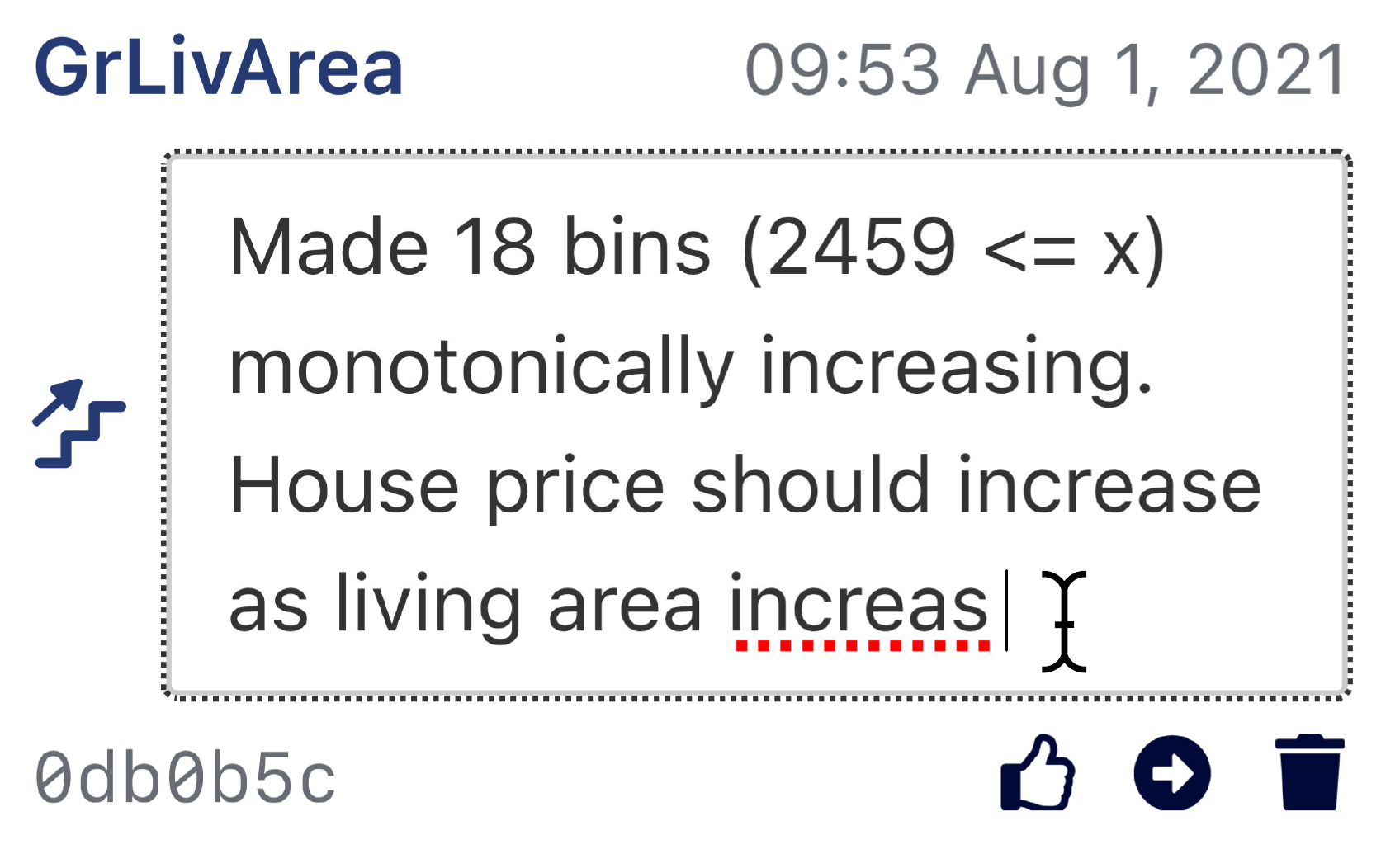}
\end{wrapfigure}

Inspired by the version control system Git\footnote{Git: \link{https://git-scm.com}}, the \historyview{}~(\autoref{fig:teaser}\figpart{-B3}) tracks each edit as a commit: a snapshot of the underlying GAM.
Each commit has a timestamp, a unique identifier, and a commit message.
Once an edit is committed, we automatically generate an initial commit message to describe the edit; users can update the message in the \historyview{} to further document their editing motivation and context.
Once users finish editing, they can click the Save button~\inlinefig{9}{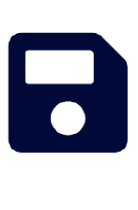} in the \statusbar{} to save the latest GAM along with all editing history, which can be used for deployment or future continuing editing.
Before saving the model, \tool{} requires users to examine and confirm~\inlinefig{9}{confirm} all edits.
 
\section{Case Studies}
\label{sec:appendix:case}

The pneumonia~(\autoref{sec:case-study-2}) and sepsis~(\autoref{sec:case-study-1}) risk prediction models are GAMs trained with boosted-trees~\cite{louAccurateIntelligibleModels2013}.
We train both models using the InterpretML library~\cite{noriInterpretMLUnifiedFramework2019}.
We use the default hyper-parameters for both models: outer bagging as \texttt{8}, inner bagging as \texttt{0}, number of interaction terms as \texttt{10}, max bins as \texttt{256}, max interaction bins as \texttt{32}, and learning rate as \texttt{0.01}.
All shape function plots of the pneumonia risk prediction model are listed in~\cite{caruanaIntelligibleModelsHealthCare2015}.

\section{User Study}
\label{appendix:user-study}

The LendingClub dataset~\cite{LendingClubOnline2018} includes 9 continuous and 9 categorical features describing the financial information of loan applicants.
The outcome variable is binary: \texttt{1} if the applicant can pay off the loan in time and \texttt{0} otherwise.
We follow the same workflow described in \autoref{sec:appendix:case} to train this model.
All shape function plots of this model are listed in~\link{https://interpret.ml/gam-changer}.
Each user study participant was given the following recommended task list and they were told that they could freely edit the model as seen fit:

\begin{itemize}[topsep=4pt, itemsep=0.7pt, parsep=0mm, leftmargin=6mm, label=$\square$]
    \item Browse different features and look for ``surprising'' patterns
    \item What are some characteristics for people with high \inlinefig{9}{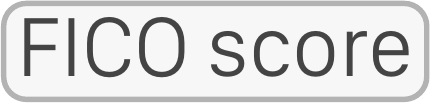}?
    \item What are some characteristics for people with low \inlinefig{9}{tag-fico}?
    \item Does these characteristics make sense?
    \item Make  \inlinefig{9}{tag-fico} shape function monotonic
    \item Increase the score for low  \inlinefig{9}{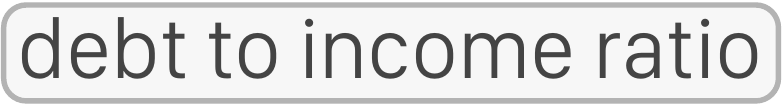}
    \item Smooth out the sudden score drop when  \inlinefig{9}{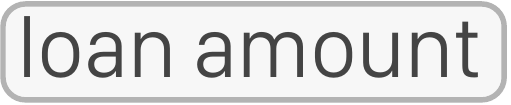} is between \texttt{\$2k} and \texttt{\$4k}
    \item Lower the score when \inlinefig{9}{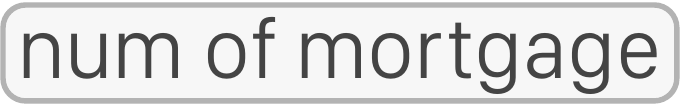} is \inlinefig{9}{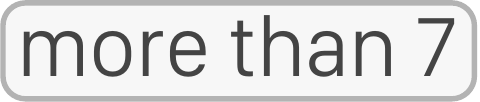}
    \item Remove the predictive effect of \inlinefig{9}{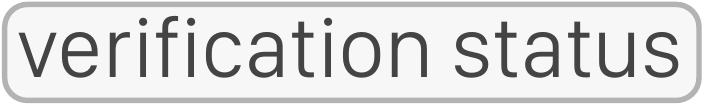} when it is  \inlinefig{9}{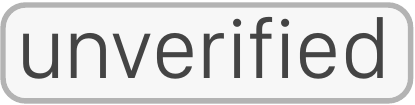}
    \item Explore the history panel
    \item Try undo and redo
    \item Try to checkout a previous edit
    \item Save the model, reload the model in a new \tool{}
    \item Free exploration
\end{itemize} 
\end{document}